%% file: main.tex
\documentclass{article} % For LaTeX2e
\usepackage{tccml_iclr2025_conference,times}

\input{math_commands.tex}

\usepackage{hyperref}
\usepackage[capitalize]{cleveref}
\usepackage{siunitx}
\usepackage{url}
\usepackage{graphicx}
\usepackage{subcaption} 
\usepackage{paralist}
\usepackage{algorithm}
\usepackage{algpseudocode}
\usepackage{wrapfig}

\title{Diffusion-LAM: Probabilistic Limited Area Weather Forecasting with Diffusion}

\author{Erik Larsson \\
Department of Computer Science\\
Linköping University, Sweden\\
\texttt{erik.larsson@liu.se} \\
\And
Joel Oskarsson \\
Department of Computer Science\\
Linköping University, Sweden\\
\texttt{joel.oskarsson@liu.se} \\
\And
Tomas Landelius \\
SMHI\\
Norrköping, Sweden\\
\texttt{tomas.landelius@smhi.se}
\And
Fredrik Lindsten \\
Department of Computer Science\\
Linköping University, Sweden\\
\texttt{fredrik.lindsten@liu.se}
}

\iclrfinalcopy % Uncomment for camera-ready version, but NOT for submission.
\begin{document}

\maketitle

\begin{abstract}
Machine learning methods have been shown to be effective for weather forecasting, based on the speed and accuracy compared to traditional numerical models. While early efforts primarily concentrated on deterministic predictions, the field has increasingly shifted toward probabilistic forecasting to better capture the forecast uncertainty. Most machine learning-based models have been designed for global-scale predictions, with only limited work targeting regional or limited area forecasting, which allows more specialized and flexible modeling for specific locations. This work introduces Diffusion-LAM, a probabilistic limited area weather model leveraging conditional diffusion. By conditioning on boundary data from surrounding regions, our approach generates forecasts within a defined area. Experimental results on the MEPS limited area dataset demonstrate the potential of Diffusion-LAM to deliver accurate probabilistic forecasts, highlighting its promise for limited-area weather prediction.
\end{abstract}

\section{Introduction}
The frequency and cost of extreme weather events appear to be increasing \citep{noaa2025billiondollar, ipcc2023synthesis, whitt2023economic}, driven by climate change \citep{ipcc2023synthesis}. Therefore, accurate and reliable weather forecasts have become increasingly crucial for a variety of downstream applications. These include early warnings for extreme weather events, optimized agricultural and food production, and efficient renewable energy planning. More efficient forecasting systems also help reduce the energy footprint of weather forecasting. In weather forecasting, ensemble forecasting is a technique used to account for uncertainty by generating multiple forecasts, where each ensemble member represents a potential future state of the atmosphere. By analyzing the full ensemble, meteorologists can quantify uncertainty and assess the likelihood of different future scenarios. More efficient forecasting systems could enable the use of larger ensembles, improving uncertainty quantification and enhancing the ability to anticipate forecast failures. We expand further on the climate change impact in relation to weather forecasting in \cref{apx:soc_impact}.

Traditionally, weather forecasting has been done with Numerical Weather Prediction (NWP), consisting of complex physical models based on differential equations running on large supercomputers \citep{NWP}. However, lately, there has been a shift to data-driven Machine Learning Weather Prediction (MLWP) due to its strong performance \citep{graphcast, pangu}. Early efforts in MLWP primarily focused on developing global deterministic models \citep{graphcast, pangu}. However, capturing the inherent uncertainty in weather predictions requires probabilistic models, and recent advancements have begun to address this need \citep{gencast,oskarsson2024probabilistic,couairon2024archesweatherarchesweathergendeterministic}. While global models show promising results, regional weather forecasting has received considerably less attention, with a few recent exceptions \citep{stretch_grid_norway, stormCast, oskarsson2023graph-lam, oskarsson2024probabilistic, xu2024yinglongskillfulhighresolution}.

\begin{figure}[H]
\begin{center}
\includegraphics[width=\textwidth]{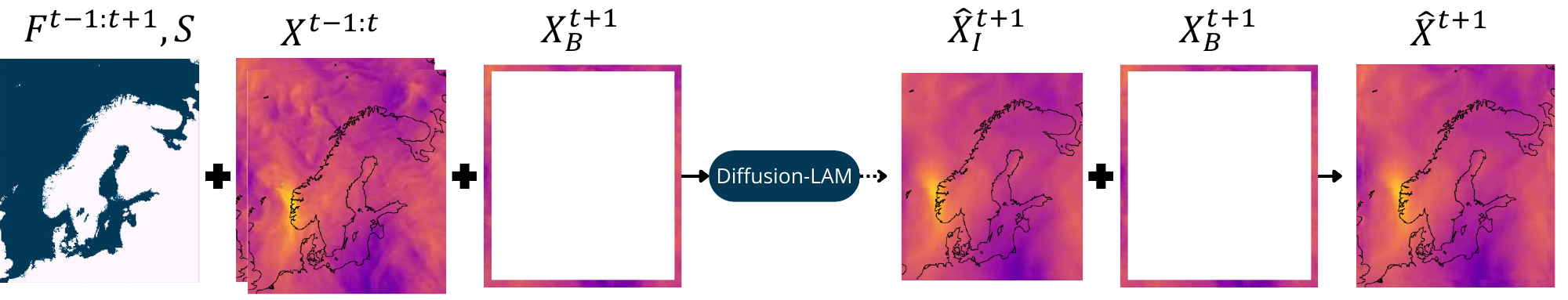}
\end{center}
\caption{An overview of the forecasting process showing the inputs and outputs of the model.}
\label{fig:model_process}
\end{figure}

\paragraph{Problem definition.}
In this paper, we tackle the problem of probabilistic MLWP Limited Area Modeling (LAM).
In LAM forecasting, the data is represented on a regular grid $G$ of dimensions ${W \times H}$ where each weather state $X^t \in \R^{G \times d_x}$ at lead time $t$ has $d_x$ variables for each position in the grid. Additionally, we have access to forcing variables $F^t$ (see \cref{tab:dataset_forcing}), which provide known quantities such as the time of day.
There are also static variables $S$, which are features associated with the grid positions such as orography and land-sea mask (see \cref{tab:dataset_static}). We divide the data into an interior input $I^t = \{X_{I}^{t-1:t}, F_{I}^{t-1:t+1}, S_{I}\}$ and a boundary input $B^t = \{X_{B}^{t-1:t+1}, F_{B}^{t-1:t+1}, S_{B}\}$ (see \cref{fig:model_process}) and define the forecasting problem as sampling from $p(X^{t+1}_{I}|I^t, B^t)$. This approach differs from previous work \citep{stormCast, xu2024yinglongskillfulhighresolution, oskarsson2023graph-lam, oskarsson2024probabilistic}, where information from the boundary or global model is only included up to the current time step $t$ as an explicit input to the forecasting model. However, incorporating also $X^{t+1}_{B}$ as an input is feasible, as it can be obtained from a global forecasting model in an operational setting. As we show in \cref{sec:results}, conditioning on $X^{t+1}_{B}$ results in forecasts that better agree with the boundary input. In practice, we learn a model that can make forecasts for a predefined forecast length, and to make longer forecasts, we roll out the model autoregressively using predicted states as input.

\paragraph{Our main contributions are:}\begin{enumerate}
    \item We propose a new framework for encoding boundary information in the LAM setting, allowing conditioning on boundary conditions from a global forecast also at future time steps.
    This results in better alignment with the boundary compared to previous methods.
    \item We develop a conditional diffusion model tailored to LAM weather forecasting making use of the boundary encoding framework.
    \item We show in experiments on the MEPS LAM dataset that the model achieves accurate ensemble forecasts with highly detailed and physically realistic fields.
\end{enumerate}

\section{Related Work} 
Ensemble MLWP can be done in multiple ways, such as perturbations to the input data \citep{chen2023fuxi, pathak2022fourcastnet, pangu, graubner2022calibration, bulte2024uncertainty} or by generative models based on latent variable formulations \citep{oskarsson2024probabilistic, SwinVRNN}, diffusion \citep{gencast, andrae2024continuousensembleweatherforecasting, shi2024codicastconditionaldiffusionmodel}, or flow-matching \citep{couairon2024archesweatherarchesweathergendeterministic}.

While directly using global MLWP forecasts (possibly with downscaling) for a specific region is possible, few works actually simulate the physics at high resolution only over a region of interest.
\citet{stretch_grid_norway} propose a stretched-grid approach to make regional forecasts with a global model. They focus on the Nordic region, where they use a higher resolution. 
The method is however deterministic and less modular and scalable, as it also necessitates learning to simulate global dynamics.
Existing deterministic LAM models include YingLong \citep{xu2024yinglongskillfulhighresolution} and Hi-LAM \citep{oskarsson2023graph-lam}.
StormCast \citep{stormCast} generates regional ensemble forecasts by deterministically predicting a mean and then applying diffusion to residuals for creating ensemble members. Unlike our work, StormCast conditions on a lower-resolution global model for the entire region rather than only the boundary. While StormCast has a high temporal resolution of \SI{1}{\hour}, their experiments are limited to forecasts up to only \SI{12}{\hour}.
\citet{oskarsson2024probabilistic} propose Graph-EFM, that produce probabilistic LAM forecasts based on a latent variable formulation. In contrast to our work, all methods above only leverage past boundary information $X^{t-1:t}_B$ up to the current time step $t$, resulting in discontinuities at the edge of the forecasting region (as we show for Graph-EFM in \cref{sec:results}). 
%We argue that this approach is suboptimal and propose a new method that integrates the boundary area of the next time step $X^{t+1}_B$ as well into each inference step, improving accuracy and consistency in limited area forecasting. 
Our method, as well as \citet{oskarsson2023graph-lam, oskarsson2024probabilistic}, extend to \SI{57}{\hour} forecasts, albeit at a temporal resolution of \SI{3}{\hour}. We explore related work in more detail in \cref{apx:related_work}.

\section{Probabilistic limited area weather forecasting with diffusion}
\begin{figure}[H]
\begin{center}
\includegraphics[width=1.0\textwidth]{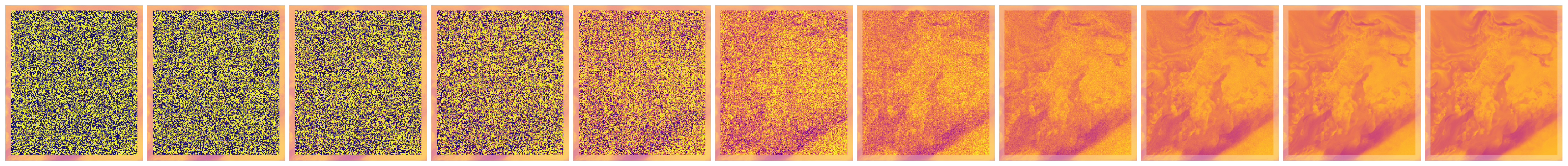}
\end{center}
\caption{The noise process for \texttt{r\_2} (relative humidity). We only show 10 diffusion steps to make the visualization simpler, but in practice use 20 steps when sampling new trajectories.}
\label{fig:noise_process}
\end{figure}

\paragraph{Conditional diffusion.}
\citet{gencast} showed that diffusion models can be a powerful tool for generating accurate probabilistic global forecasts. We therefore design our model using the same diffusion framework, originating from \citet{karras2022elucidating}. The denoising diffusion model starts with sampling a latent noise variable $Z_0^t \sim \mathcal{N}(0, \sigma^2_0 \mI)$ and iteratively denoise $Z_n^t, n \in \{0, 1, ..., N\}$ for $N$ steps until we reach the data distribution at $Z_N^t$, as visualized in \cref{fig:noise_process}. In practice, we don't predict $X^{t+1}_I$, but the residual $(X^{t+1}_I - X^{t}_I)$, which we then add to the current state $X^t_I$. Each step in the denoising process is conditioned on $I^t, B^t$, which can be interpreted as a conditional inpainting task. We describe conditional diffusion in more detail in \cref{apx:model_details}.
\paragraph{Model.}
Building on the conditional diffusion framework described above, we design a model architecture that incorporates \(\{I^t, B^t\}\) as conditioning inputs throughout the denoising process. Since $Z_n^t$ and $I^t$ have the same spatial dimensions, we concatenate the tensors along the feature dimension. We encode the grid using two separate pixel-wise MLPs with 1 hidden layer, one for the interior $\text{MLP}_I(\{I^t, Z^t_n\})$ and the other for the boundary $\text{MLP}_B(B^t)$. The boundary encoder operates exclusively on the boundary, while the interior encoder processes only the interior grid positions. This is more suitable when denoising the interior but not the boundary. After encoding the interior and the boundary separately, we re-assemble the full regular grid by combining the encoded interior and the boundary to get a $W \times H \times C$ feature tensor, which we then pass to a U-Net \citep{UNET_Original}. 
We use the U-Net architecture due to its high efficiency for data on a regular grid \citep{siddiqui2024exploringdesignspacedeeplearningbased}.
Our U-Net is adapted from \citet{song2020score, karras2022elucidating} with adaptive padding to enable arbitrary grid shapes. The diffusion noise is encoded with Fourier embeddings similarly to \citet{karras2022elucidating} and added to the network through conditional normalization layers. Our model is trained on making \SI{3}{\hour} forecasts (the process is shown in \cref{fig:model_process}), but to make longer forecasts, we roll out the model autoregressive using predicted states as input. A forecast trajectory of length $T$ steps can then be defined as
$$
p\left(X_{I}^{1:T}\middle|I^{0:T-1}, B^{0:T-1}\right) = {\textstyle \prod^{T-1}_{t=0}} p\left(X_{I}^{t+1}\middle|I^t, B^t\right).
$$
Further model details are described in \cref{apx:model_details}.

\paragraph{Training.}
During training, we apply noise to each residual from a uniformly sampled noise level $n$. A single denoising step is then performed to make a prediction $\hat{X}^{t+1}_I$ of the next state $X^{t+1}_I$ before computing the training loss. The loss function 
\begin{equation}\label{eq:training_loss}
   \mathcal{L_{\text{WMSE}}} = \mathbb{E}_{n \sim \text{Uniform}(0, N-1)} \left[\frac{1}{|G_I|} {\textstyle \sum_{g \in G_{I}} \sum^{d_x}_{d=1}} h_l \lambda_d \omega_n \left(\hat{X}^{t+1}_{g, d} - X^{t+1}_{g, d}\right)^2\right] 
\end{equation}
is a weighted MSE denoising loss with $G_I$ the set of interior grid points.
The loss includes three scaling components: for atmospheric level $h_l$, per variable $\lambda_d$, and for the diffusion noise level $\omega_n$. Unlike \citet{oskarsson2024probabilistic}, we do not perform autoregressive training over multiple steps, as sampling forecasts during training is computationally prohibitive. Nevertheless, our model demonstrates comparable stability without autoregressive training, consistent with observations in other diffusion-based methods \citep{gencast,stormCast}. 
Training only on a single time step simplifies the training process and significantly reduces GPU memory requirements, potentially allowing for higher-resolution inputs.

\section{Experiments}
To evaluate our model, we conduct experiments on LAM forecasting using the MEPS dataset\footnote{The MEPS dataset is openly available at \url{https://nextcloud.liu.se/s/meps}} and measure root mean squared error (RMSE), continuous ranked probability score (CRPS), and spread-skill ratio (SSR). The metric computations are explained in \cref{apx:metrics}. 
The MEPS dataset contains NWP forecasts for the Nordic region from the MetCoOp Ensemble Prediction System. Since we are training on forecasts, the objective is not to outperform MEPS but rather to develop a more efficient emulator model that could for example be used to create larger ensembles. The dataset consists of \SI{6069} forecasts represented on a $238 \times 268$ grid with \SI{10}{\kilo\meter} spatial resolution and a temporal resolution of \SI{3}{\hour}, up to a maximum lead time of \SI{57}{\hour}. Each grid point includes 17 atmospheric fields at various heights and pressure levels, as well as static and forcing features. The outermost 10 grid points define the boundary region, which, in operational settings, could be provided by a re-gridded forecast from a global model. Further dataset details are given in \cref{apx:dataset_details}.

We sample \SI{57}{\hour} forecasts with 25-ensemble members using batched sampling in \SI{8}{\minute} (\SI{20}{\second} per ensemble member) on a single \SI{80}{\giga\byte} A100 GPU. Compared to deterministic or latent variable models, diffusion models require more time to generate forecasts due to the need for multiple forward passes. However, they remain relatively efficient when compared to traditional NWP models. We compare Diffusion-LAM (5 and 25 ensemble members) to Graph-EFM \citep{oskarsson2024probabilistic} (5 and 25 members), as it is the approach most similar to ours. 
It is probabilistic, conditions only on the boundary rather than the entire domain (unlike StormCast \citep{stormCast}), and provide publicly available code for both training and inference. 
Additionally, we compare to a version of Diffusion-LAM without the boundary conditioning on the next time step $X^{t+1}_B$ (no boundary).

\subsection{Results}\label{sec:results}
\begin{figure}[H]
    \centering
        \includegraphics[width=0.9\textwidth]{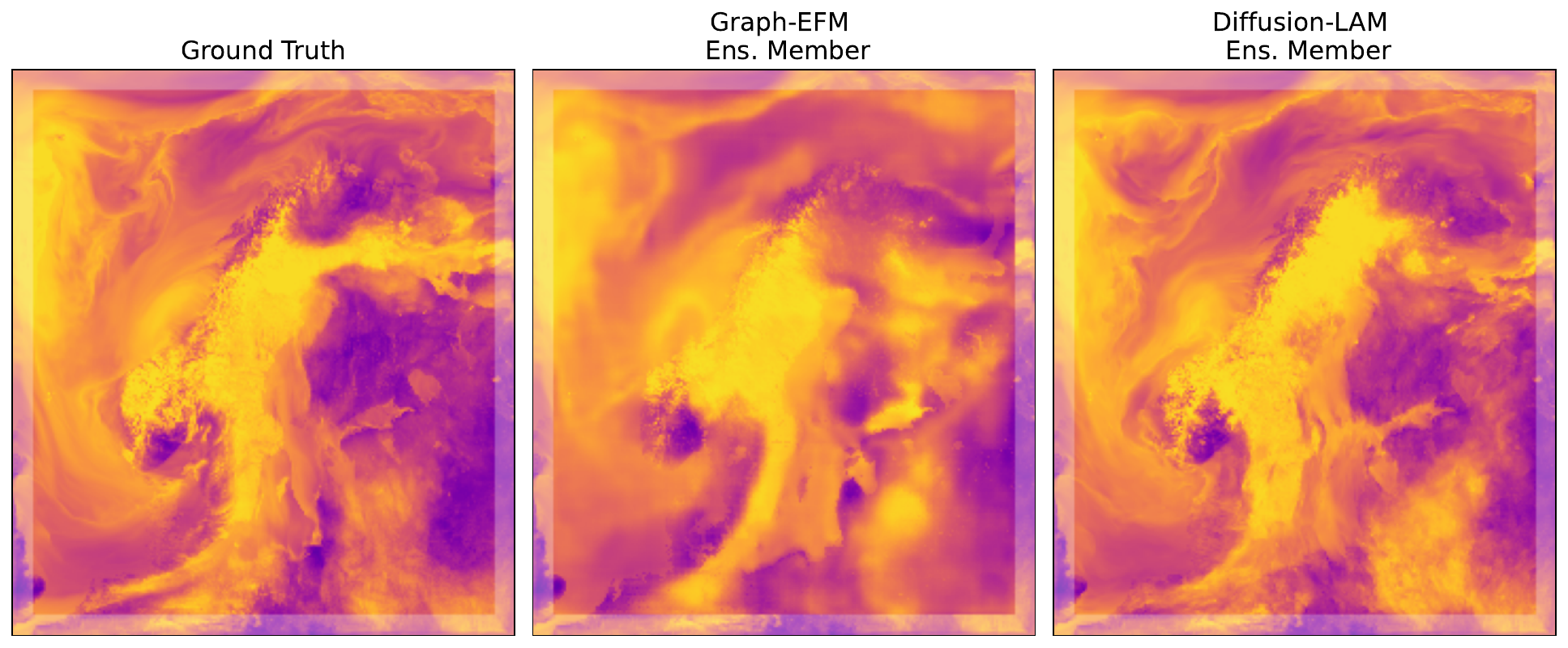}
    \caption{Forecasts at \SI{57}{h} lead time for \texttt{r\_2}. The faded area constitutes the boundary region. Note the difference in fine-scale details and the consistency with the boundary in the ensemble members.}
    \label{fig:qualitative_results}
\end{figure}

\begin{figure}[H]
    \centering
    \includegraphics[width=\textwidth]{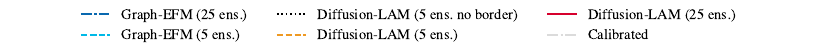}
    \begin{subfigure}[b]{0.3\textwidth}
        \centering
        \includegraphics[width=\textwidth]{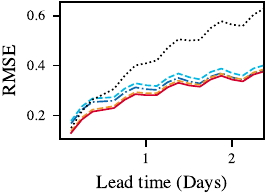}
        \label{fig:subfigure1}
    \end{subfigure}
    \hfill
    \begin{subfigure}[b]{0.3\textwidth}
        \centering
        \includegraphics[width=\textwidth]{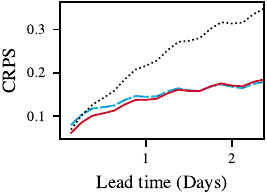}
        \label{fig:subfigure1}
    \end{subfigure}
    \hfill
    \begin{subfigure}[b]{0.3\textwidth}
        \centering
        \includegraphics[width=\textwidth]{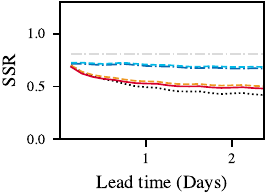}
        \label{fig:subfigure2}
    \end{subfigure}
    \caption{The mean of the normalized RMSE, CRPS, and SSR for all variables.}
    \label{fig:quantitative_results}
\end{figure}

The forecasts in \cref{fig:qualitative_results} show that Diffusion-LAM can produce much more realistic and less smooth ensemble members than Graph-EFM. We also observe that our model demonstrates significantly better consistency with the boundary conditions, while Graph-EFM occasionally deviates significantly from patterns on the boundary (see for example the bottom right corner of the forecasts in \cref{fig:qualitative_results}).
As can be seen in \cref{fig:quantitative_results} our model outperforms Graph-EFM in terms of RMSE and CRPS for shorter lead times. However, at longer lead times performance is similar. 
While Diffusion-LAM (no border) has a comparable error for single step predictions, it grows quickly, emphasizing the importance of including information from $X^{t+1}_B$ for accurate roll-outs with the diffusion model. 
Both models struggle to generate an adequate spread ($\text{SSR} \approx 1$), indicating that the uncertainty captured by the model is somewhat underestimated. The SSR is comparable for single-step predictions, but Diffusion-LAM struggles to maintain sufficient ensemble spread at longer lead times, suggesting a potential issue with the roll-out procedure.
The difference between using 5 and 25 ensemble members is small for both models.
Detailed results for all variables are available in \cref{apx:detailed_results}.

\section{Conclusion}
This work introduces a new framework for integrating boundary conditions from global forecasts of the next time step in the prediction step. We present Diffusion-LAM, a probabilistic MLWP LAM model that leverages an improved framework for integrating boundary conditions also from future time steps. By experiments on the MEPS LAM dataset we demonstrate that our method delivers accurate ensemble forecasts with much more detailed and physically realistic fields. Promising directions for future research include exploring more realistic scenarios that better reflect operational settings, as well as enhancing the sampling speed, spread, and accuracy of diffusion MLWP models. Potential research avenues are discussed in greater detail in \cref{apx:future_work}.

\subsubsection*{Acknowledgments}
This research is financially supported by the Swedish Research Council (grant no: 2020-04122, 2024-05011)
the Wallenberg AI, Autonomous Systems and Software Program (WASP) funded by the Knut and Alice Wallenberg Foundation,
and
the Excellence Center at Linköping--Lund in Information Technology (ELLIIT).
Our computations were enabled by the Berzelius resource at the National Supercomputer Centre, provided by the  Knut and Alice Wallenberg Foundation.
Landelius was funded by the Swedish Energy Agency and the European Union’s Horizon 2020 research and innovation programme under grant agreement no. 883973–EnerDigit.

\bibliography{iclr2025_conference}
\bibliographystyle{iclr2025_conference}

\clearpage% Added this for clear separation
\appendix
\section{Table of notation}
The notation that is used in this paper is summarized in \cref{tab:notation}.
\begin{table}[h]
\caption{Table of notation.}
\label{tab:notation}
\begin{center}
\begin{tabular}{ll}
\multicolumn{1}{l}{\bf Notation}  &\multicolumn{1}{l}{\bf Description}
\\ \hline \\
$X^t$ & Full weather state including both the interior and the boundary at lead time $t$ \\
$X^t_{I}$ & Interior of weather state at lead time $t$ \\
$X^t_{B}$ & Boundary of weather state at lead time $t$ \\
$X^t_{g, d}$ & Weather variable $d$ at grid position $g$ at lead time $t$ \\
$\hat{X}^t_{g, d}$ & Predicted weather variable $d$ at grid position $g$ at lead time $t$ \\
$G$ & The grid dimension of a full weather state $X^t$ \\
$W$ & The width of the regular grid $G$ \\
$H$ & The height of the regular grid $G$ \\
$C$ & The number of feature channels for the encoded grid \\
$G_{I}$ & The grid dimension of the interior of a weather state $X^t_{I}$ \\
$F^t$ & Forcing variables at lead time $t$ \\
$S$ & Static variables for each position in the grid $G$ \\
$I^t$ & Interior input $\{X_{I}^{t-1:t}, F_{I}^{t-1:t+1}, S_{I}\}$ at lead time $t$ \\
$B^t$ & Boundary input $\{X^{t-1:t+1}_{B}, F^{t-1:t+1}_{B}, S_{B}\}$ at lead time $t$ \\
$d_x$ & The number of weather variables in each grid cell of each state $X^t$ \\
$T$ & Number of forecast steps in a sampled trajectory \\
$N_{ens}$ & Number of ensemble members \\
$Z^t_n$ & Latent noise at lead time $t$ and noise level $n$ \\
$h_l$ & The weight for the loss function for height/pressure level $l$\\
$\lambda_d$ & The variable weight for the loss function for variable $d$ \\
$\omega_n$ & The weight for the loss function at noise level $n$ \\
$\sigma_n$ & The noise at noise level $n$ \\
$\sigma_{data}$ & The expected standard deviation of the data \\
$\sigma_{min}$ & The minimum noise level for the diffusion process \\
$\sigma_{max}$ & The maximum noise level for the diffusion process \\
\end{tabular}
\end{center}
\end{table}

\section{Societal impact}\label{apx:soc_impact}
Here we expand further on the societal impact of weather forecasting in relation to extreme weather, forecast failures, agriculture and food, renewable energy, and the energy footprint of weather forecasting.

\subsection{Extreme weather}
Extreme weather events, including droughts, floods, freezes, severe storms, tropical cyclones, wildfires, and winter storms, can cause over 50 billion US dollars in damages annually in the United States \citep{noaa2025billiondollar}. As shown in \cref{tab:extreme_weather}, the annual number of events and total damages have increased more than eightfold from the 1980s to 2024. Although the increase in the number of deaths is smaller, the number of deaths has nearly doubled over the same period.

\begin{table}[h]
\caption{The consequences of extreme weather in the United States \citep{noaa2025billiondollar}. The cost is in billion US dollars.}
\label{tab:extreme_weather}
\begin{center}
\begin{tabular}{llll}
\multicolumn{1}{c}{\bf Time Period}  &\multicolumn{1}{c}{\bf Events/Year} & \multicolumn{1}{c}{\bf Cost/Year} & \multicolumn{1}{c}{\bf Deaths/Year}
\\ \hline \\
1980s (1980-1989) & 3.3 & 22.0 & 299 \\
1990s (1990-1999) & 5.7 & 33.5 & 308 \\
2000s (2000-2009) & 6.7 & 62.1 & 310 \\
2010s (2010-2019) & 13.1 & 99.5 & 523 \\
Last 5 Years (2020-2024) & 23.0 & 149.3 & 504 \\
Last 3 Years (2022-2024) & 24.3 & 153.9 & 511 \\
Last Year (2024) & 27.0 & 182.7 & 568 \\
\end{tabular}
\end{center}
\end{table}

Alarmingly, there has been a clear upward trend in the frequency, cost, and associated deaths since the 1980s. This trend is believed to affect the entire globe and to be driven by climate change \citep{ipcc2023synthesis, whitt2023economic}. Since all regions of the globe are impacted, the development of cost-effective forecasting systems can significantly enhance the accessibility of accurate forecasts and early warning systems for the world’s most economically and socially vulnerable populations, which are often disproportionately affected by disasters \citep{worldbank2023inclusive}. Given the substantial economic and human costs of extreme weather, improving the accuracy of weather forecasting is increasingly critical to mitigating its impacts. 

\subsection{Forecast Failures}  
Forecast errors are an inevitable challenge in weather prediction systems due to limitations in model design and data availability \citep{leutbecher2008ensemble, yano2018scientific}. Even with advanced models, perfect representation of atmospheric dynamics is unattainable because of factors such as incomplete observations, resolution constraints, and inherent chaos in weather systems.  

In traditional physical models, the governing equations and physical assumptions provide a clear foundation \citep{bauer2015quiet}, making it easier to diagnose and understand the causes of forecast errors. In contrast, MLWP models, which rely on data-driven approaches, often behave as black boxes. The complexity of these models can make it difficult to interpret why a forecast fails or how errors propagate, posing significant challenges for transparency and trust in critical applications.  

To address these uncertainties, probabilistic forecasts are increasingly used to provide a range of possible outcomes rather than a single deterministic prediction. This approach enables uncertainty quantification, allowing users to make more informed decisions by understanding the likelihood of various weather scenarios.

\subsection{Agriculture and food}
Reliable weather prediction systems are crucial for the agriculture and food sectors, as they directly influence food security and economic stability \citep{ipcc2023synthesis}. Agriculture is highly sensitive to weather conditions \citep{whitt2023economic}, and the ability to anticipate weather patterns plays a key role in ensuring efficient and productive farming practices. By having access to early warnings, farmers can take preventive measures to protect crops, such as using irrigation systems during droughts, using frost protection techniques, or securing infrastructure during storms.  

\subsection{Renewable energy}
Renewable energy sources are inherently variable and highly dependent on current weather conditions, making accurate weather forecasting crucial for predicting future energy generation and ensuring system stability \citep{ren_energy_future, SHARMA2014160, ren_energy_pred}. In the short term, precise forecasts can facilitate an efficient integration of renewable energy into existing power grids \citep{ren_energy_future}. Additionally, advancing weather forecasting over longer timescales, such as seasonal or climate modeling, can support more effective planning and decision-making for renewable energy systems \citep{ren_energy_future}.

\subsection{The energy footprint of weather forecasting}
Current NWP models require significant computational resources to generate forecasts \citep{fundamentals_of_nwp, bauer2015quiet}, leading to a high energy footprint. MLWP models, on the other hand, can be far more efficient during inference. However, the training cost of MLWP systems must also be considered. For instance, training FourCastNet consumes a similar amount of energy as running a single 10-day forecast with 50 ensemble members using traditional NWP \citep{pathak2022fourcastnet}, and these models are unlikely to require retraining every 10 days. 

Reducing the computational resources needed to produce one ensemble forecast does not automatically translate to lower energy consumption, as the saved resources could instead be allocated to generating a larger number of ensembles. This creates a trade-off where we can either produce the same number of ensembles more quickly and at a lower cost, or utilize the same budget to significantly increase the number of ensembles.

\section{Related work}\label{apx:related_work}
Many MLWP models have been developed for global deterministic weather forecasting utilizing various architectures \citep{siddiqui2024exploringdesignspacedeeplearningbased}. The architecture choices include fixed-grid frameworks such as convolutional neural networks \citep{unet_weather} and transformer-based models \citep{pangu, couairon2024archesweatherarchesweathergendeterministic}. Grid-invariant approaches have also gained traction, utilizing graph-based architectures \citep{keisler2022forecastingglobalweathergraph, graphcast, oskarsson2024probabilistic, lang2024aifsecmwfsdatadriven} and operator-based methods \citep{pathak2022fourcastnet, bonev2023sfno}. Additionally, hybrid models that integrate NWP with MLWP have been explored \citep{neuralGCM, verma2024climode}. Recent efforts have even explored shifting away from grid-based representations of the data, focusing exclusively on learning directly from observations \citep{mcnally2024datadrivenweatherforecasts}. However, since the data in our LAM formulation is represented on a regular grid, we follow the recommendations of \citet{siddiqui2024exploringdesignspacedeeplearningbased} and adopt a U-Net architecture.

%\citet{xu2024yinglongskillfulhighresolution} presents YingLong, a deterministic LAM that uses a boundary smoothing technique where the prediction on the boundary area is a combination of the interior MLWP model and the boundary conditions from a NWP model. Their method combines a swin transformer \citep{liu2021swin} and adaptive Fourier neural operators \citep{AFNO} to capture the local and global features, respectively. \citet{oskarsson2023graph-lam} introduces a deterministic hierarchical graph neural network for LAM and 

\section{Dataset details}\label{apx:dataset_details}
Since the training objective is based on forecasts rather than actual observations, the objective is to develop an emulator model for MEPS. The \SI{6069} forecasts in the dataset are from the time period April 2021 to March 2023. For simplicity and consistency with \citet{oskarsson2024probabilistic} we use the same training, validation and test split. We use forecasts from April 2021 to June 2022 for training (\SI{2713} samples) and validation (\SI{678} samples), and forecasts from July 2022 to March 2023 for testing (\SI{2678} samples).

\begin{table}[h!]
\caption{Variables in the MEPS dataset. \textsuperscript{*}Level 65 in the MEPS system is approximately \SI{12.5}{\meter} over the ground \citep{arome_metcoop}.}
\label{tab:dataset_variables}
\begin{center}
\begin{tabular}{lccc}
\multicolumn{1}{l}{\bf Description}  & \multicolumn{1}{c}{\bf Abbreviation} & \multicolumn{1}{c}{\bf Unit} & \multicolumn{1}{c}{\bf Residual standard deviation}
\\ \hline \\
Net longwave solar radiation flux at the surface & nlwrs & \si{\watt\per\metre\squared} & $0.0583$ \\
Net shortwave solar radiation flux at the surface & nswrs & \si{\watt\per\metre\squared} & $0.0583$ \\
Atmospheric pressure at ground level & pres\_0g & \si{\pascal} & $0.6399$ \\
Atmospheric pressure at sea level & pres\_0s & \si{\pascal} & $0.7608$ \\
Relative humidity at \SI{2}{\meter} & r\_2 & [0, 1] & $0.5534$ \\
Relative humidity at level 65\textsuperscript{*} & r\_65 & [0, 1] & $0.5371$ \\
Temperature at \SI{2}{\meter} & t\_2 & \si{\kelvin} & $0.2197$ \\
Temperature at level 65\textsuperscript{*} & t\_65 & \si{\kelvin} & $0.1950$ \\
Temperature at \SI{500}{\hecto\pascal} & t\_500 & \si{\kelvin} & $0.1319$ \\
Temperature at \SI{850}{\hecto\pascal} & t\_850 & \si{\kelvin} & $0.1294$ \\
$u$-component of wind at level 65\textsuperscript{*} & u\_65 & \si{\meter\per\second} & $0.3885$ \\
$u$-component of wind at \SI{850}{\hecto\pascal} & u\_850 & \si{\meter\per\second} & $0.3530$ \\
$v$-component of wind at level 65\textsuperscript{*} & v\_65 & \si{\meter\per\second} & $0.3815$ \\
$v$-component of wind at \SI{850}{\hecto\pascal} & v\_850 & \si{\meter\per\second} & $0.3861$ \\
Water vapor for the full integrated column & wvint\_0 & \si{\kilo\gram\per\meter\squared} & $0.2473$ \\
Geopotential at \SI{1000}{\hecto\pascal} & z\_1000 & \si{\meter\squared\per\second\squared} & $0.1202$ \\
Geopotential at \SI{500}{\hecto\pascal} & z\_500 & \si{\meter\squared\per\second\squared} & $0.0720$ \\
\end{tabular}
\end{center}
\end{table}

\begin{table}[h!]
\caption{Forcing features in the MEPS dataset.}
\label{tab:dataset_forcing}
\begin{center}
\begin{tabular}{lcc}
\multicolumn{1}{l}{\bf Description}  & \multicolumn{1}{c}{\bf Abbreviation} & \multicolumn{1}{c}{\bf Unit}
\\ \hline \\
Solar radiation flux at the top of the atmosphere & toa & \si{\watt\per\meter\squared} \\
Fraction of open water at the surface & water & [0, 1] \\
Sine-encoded time of day & sin\_tod & [0, 1] \\
Cosine-encoded time of day & cos\_tod & [0, 1] \\
Sine-encoded time of year & sin\_toy & [0, 1] \\
Cosine-encoded time of year & cos\_toy & [0, 1] \\
\end{tabular}
\end{center}
\end{table}

\begin{table}[h!]
\caption{Static features for each grid position in the MEPS dataset.}
\label{tab:dataset_static}
\begin{center}
\begin{tabular}{lcc}
\multicolumn{1}{l}{\bf Description}  & \multicolumn{1}{c}{\bf Abbreviation} & \multicolumn{1}{c}{\bf Unit}
\\ \hline \\
Topology (geopotential at the surface) & topology & \si{\meter\squared\per\second\squared} \\
x-coordinate in the MEPS projection & x\_coord & [0, 1] \\
y-coordinate in the MEPS projection & y\_coord & [0, 1] \\
Boundary mask (indicating which pixels belong to the border) & border\_mask & 0/1 \\
Interior mask (indicating which pixels belong to the interior) & interior\_mask & 0/1 \\
\end{tabular}
\end{center}
\end{table}

\section{Model details}\label{apx:model_details}
Here, we provide additional details about Diffusion-LAM. Following common practice in MLWP models \citep{graphcast, gencast, oskarsson2023graph-lam, oskarsson2024probabilistic, couairon2024archesweatherarchesweathergendeterministic}, we use both the current and previous states as initial conditions when predicting the next state, rather than relying solely on the current state. This approach allows the model to capture first-order state dynamics more effectively.

\paragraph{Conditional diffusion.} In the diffusion process, we want to go from the initial noisy sample $Z_0^t$ to $Z_N^t$. This is achieved by using an ODE solver to the probability flow ODE
$$
\delta x = - \dot{\sigma}(t)\sigma(t)\nabla_x \log p(x;\sigma(t)) \mathrm{d}t.
$$
Each step in this solver is denoted by $D_{\theta}$ with
$$
Z_{n+1}^t = D_{\theta}(Z_n^t, I^t, B^t, \sigma_{n+1}, \sigma_n), \quad n \in 0, 1, 2,..., N,
$$
taking us from a noise level $\sigma_n$ to $\sigma_{n+1} < \sigma_n$, conditioned on $\{I^t, B^t\}$. In practice $D_{\theta}$ is parametrized with another network $F_{\theta}$ by
$$
D_{\theta}(Z_n^t, I^t, B^t, \sigma_{n+1}, \sigma_n) = c_{\text{skip}}(\sigma_n) \cdot Z_n^t + c_{\text{out}}(\sigma_n) \cdot F_{\theta}(c_{\text{in}}(\sigma_n) \cdot Z_n^t, c_{\text{noise}} (\sigma_n), I^t, B^t),
$$
where
$$
c_{\text{skip}}(\sigma_n) = \frac{\sigma_{\text{data}}^2}{\sigma_n^2 + \sigma_{\text{data}}^2}
$$
$$
c_{\text{out}}(\sigma_n) = \frac{\sigma_n^2 \cdot \sigma_{\text{data}}^2}{\sqrt{\sigma_n^2 + \sigma_{\text{data}}^2}}
$$
$$
c_{\text{in}}(\sigma_n) = \frac{1}{\sqrt{\sigma_n^2 + \sigma_{\text{data}}^2}}
$$
$$
c_{\text{noise}} (\sigma_n) = \frac{1}{4} \ln(\sigma_n)
$$
to allow for the preconditioning as in \citet{karras2022elucidating}. The noise schedule follows
$$
\sigma_n = (\sigma_{\text{max}}^{\frac{1}{\rho}} + \frac{n}{N-1}(\sigma_{\text{min}}^{\frac{1}{\rho}} - \sigma_{\text{max}}^{\frac{1}{\rho}}))^{\rho}, \quad \sigma_N = 0.
$$
During the sampling process we use a 2nd order Heun solver and take $N=20$ solver steps with $n \in \{0, 1, ...,  N-1\}$ per generated forecast. Since we are using a second-order solver this results in $N \times 2 - 1 = 39$ sequential forward passes with $D_{\theta}$. To generate ensemble forecasts we can simply sample a new $Z_0^t \sim \mathcal{N}(0, \sigma^2_0 \mI)$ for each ensemble member.

\begin{wrapfigure}[19]{r}{0.3\textwidth} % 'r' for right, and 0.4\textwidth for the figure width
    \centering
    \includegraphics[width=\linewidth]{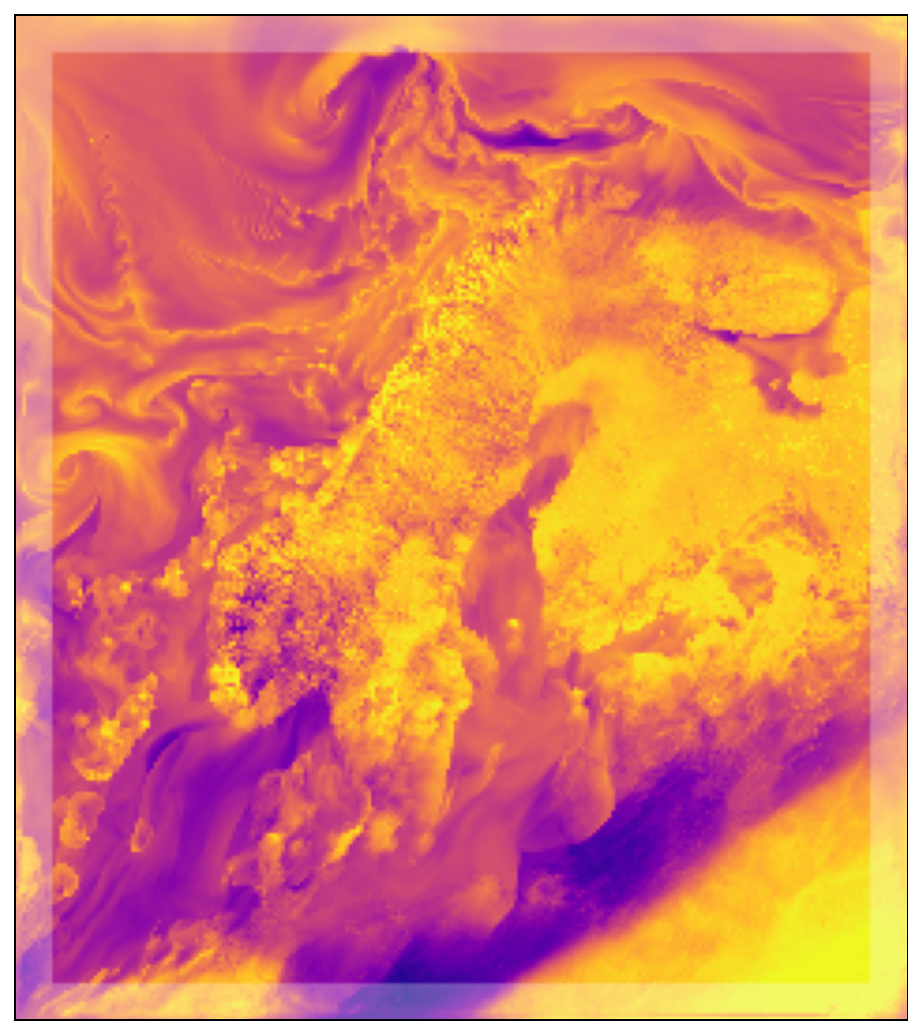}
    \caption{The interior and boundary of a weather state in our limited area model. The faded area is the 10 outermost grid positions, which we use as the boundary area.}
    \label{fig:border}
\end{wrapfigure}

\paragraph{Model.} The architecture of the model follows the encode, process, decode framework. An overview of the prediction process is shown in \cref{fig:model_process} and a visualization of the denoising process is shown in \cref{fig:noise_process}. Firstly, we encode the grid using a separate MLP encoder for the interior and the boundary. The boundary encoder operates exclusively on the faded pixels in \cref{fig:border}, while the interior encoder processes only the non-faded pixels. The encoder consists of a 1-hidden-layer MLP that acts on the feature dimension and pixel-wise maps the input to a latent space of dimension $128$. We then re-assemble the grid by combining the interior $\{\text{MLP}_I(I^t), \text{MLP}_I(Z^t_n)\}$ and the boundary $\text{MLP}_B(B^t)$, where we combine $\text{MLP}_B(X_{B}^{t+1})$ and $\text{MLP}_I(Z^t_0)$ as the interior and boundary respectively to create an encoded feature tensor of shape $238 \times 268 \times 71$ The encoded data is then sent to the diffusion backbone which encodes the data back to the grid dimensions in the last step. 

In line with \citet{siddiqui2024exploringdesignspacedeeplearningbased} we chose the U-Net \citep{UNET_Original} architecture for the diffusion backbone due to its high efficiency, and since the data is on a regular grid, a grid-invariant architecture like graph neural networks is unnecessary.  Moreover, preliminary experiments indicated that graph neural networks were significantly slower, making them impractical for our diffusion model, which requires 39 forward passes per sample.

We make minor adaptations to the U-Net used in \citet{song2020score, karras2022elucidating} to include padding to allow for arbitrary input grid shapes. The U-Net has 128 feature channels for the top level and 256 for levels 2-4. Note that the model only makes predictions on the interior of the grid as the boundary $X^{t+1}_B$ is provided as an input. The diffusion noise is encoded with Fourier embeddings as in \citet{karras2022elucidating} by transforming the noise into a vector of since/cosine features at 32 frequencies with base period 16. The features are then passed through a 2-layer MLP with SiLU \citep{hendrycks2023gaussianerrorlinearunits} activation which results in a 512 dimensional encoding of the noise. This encoding is then added to the network through conditional layer norms in the MLP encoder and the group norms of the U-Net. The full model has \SI{63.8} million parameters. 

\section{Training details}\label{apx:training_details}
The models are trained using 1 to 8 GPUs in a data-parallel configuration. The hyperparameters used for training can be found in \cref{tab:hyperparams} and we follow the training schedule from \cref{tab:training-schedule}. We use the AdamW optimizer \citep{adamw} with $\beta_1 = 0.9$, $\beta_2 = 0.95$, and a weight decay of 0.1. 

\begin{table}[h!]
\begin{minipage}[t]{0.48\textwidth} % Left table, width = 48% of the text width
\caption{Training hyperparameters}
\label{tab:hyperparams}
\begin{center}
\begin{tabular}{ll}
\multicolumn{1}{c}{\bf Hyperparameter}  &\multicolumn{1}{c}{\bf Value} \\
\hline \\
$\sigma_{\text{max}}$ & $88$ \\
$\sigma_{\text{min}}$ & $0.02$ \\
$\rho$ & $7$ \\
\end{tabular}
\end{center}
\end{minipage}
\hfill
\begin{minipage}[t]{0.48\textwidth} % Right table, width = 48% of the text width
\caption{Training schedule}
\label{tab:training-schedule}
\begin{center}
\begin{tabular}{ll}
\multicolumn{1}{c}{\bf Epochs}  &\multicolumn{1}{c}{\bf Learning Rate} \\
\hline \\
$600$ & $0.001$ \\
$400$ & $0.0001$ \\
$200$ & $0.00001$ \\
\end{tabular}
\end{center}
\end{minipage}
\end{table}

We normalize the data by the mean and standard deviation of the training set. Then we calculate the mean and standard deviation of the residuals of the standardized training dataset. Since our target is normalized we set $\sigma_{\text{data}} = 1$. During training, we uniformly sample the noise level $n$ for each sample, add the noise $\mathcal{N}(0, \sigma^2_n \mI)$ to the target residual, and perform one denoising step before calculating the training loss.

Following \citet{graphcast, gencast, oskarsson2024probabilistic}, we weight the loss \cref{eq:training_loss} by $h_l$ for atmospheric level $l$, with detailed values provided in \cref{tab:level_weights}. This prioritizes surface variables, which are more relevant in LAMs, while down-weighting upper-atmosphere fields, where global dynamics dominate. As in \citet{oskarsson2024probabilistic}, we scale the loss by the residual standard deviation $\lambda_d$ for variable $d$ to account for fields with greater variability, which are typically harder to predict. The standard deviation for the residuals are presented in \cref{tab:dataset_variables}. Note, that the data is normalized before we compute the residuals. Inspired by \citet{karras2022elucidating}, we adjust the loss by
$$
\omega_n = \frac{\sigma_n^2 + \sigma_{\text{data}}^2}{(\sigma_n \cdot \sigma_{\text{data}})^2}
$$
for the noise level $n$ that added to the ground truth during training. Early in the diffusion process, MSE losses are higher, so scaling the loss so that higher noise samples gets a lower weight makes sure that as much emphasis is placed on the final denoising steps where predictions converge toward the ground truth.

\begin{table}[h!]
\caption{Height and pressure level weighting}
\label{tab:level_weights}
\begin{center}
\begin{tabular}{ll}
\multicolumn{1}{c}{\bf Height/Preassure}  &\multicolumn{1}{c}{\bf Weight}
\\ \hline \\
\SI{2}{\meter} & $1.0$ \\
Surface variables & $0.1$ \\
Level 65 & $0.065$ \\
\SI{1000}{\hecto\pascal} & $0.1$ \\
\SI{850}{\hecto\pascal} & $0.05$ \\
\SI{500}{\hecto\pascal} & $0.03$ \\
\end{tabular}
\end{center}
\end{table}

% Nevertheless, our model demonstrates comparable stability without autoregressive training, consistent with observations in other diffusion-based methods \citep{gencast,stormCast}. %This contrasts with many deterministic models, where autoregressive training is often essential for stability \citep{siddiqui2024exploringdesignspacedeeplearningbased}. Training only on a single time step simplifies the training process and significantly reduces GPU memory requirements, potentially allowing for higher-resolution inputs.

\section{Experiment details}\label{apx:experiment_details}
We use up to 8 \SI{80}{\giga\byte} A100 GPUs in parallel to sample trajectories for the entire test set faster. The hyperparameters used for sampling can be found in \cref{tab:inference_hyperparams}. Due to the high computational cost, we are not able to re-trainin multiple models for an extensive statistical analysis. The models\footnote{The code and implementation details will be made publicly available upon acceptance of this paper. \url{https://github.com/ErikLarssonDev/Diffusion-LAM/blob/main}} are implemented in PyTorch\footnote{\url{https://pytorch.org/}} and the code base is based on the neural lam\footnote{\url{https://github.com/mllam/neural-lam}} project.

We compare only to Graph-EFM, as it is the most similar to our approach. It is probabilistic (unlike Graph-FM and YingLong), conditions only on the boundary rather than the entire domain from a global model (unlike StormCast \citep{stormCast}), and includes experiments on the MEPS dataset. Additionally, Graph-EFM is the only probabilistic LAM model we know of with publicly available code for both training and inference. We sample forecasts from Graph-EFM using the original configuration as described by \citet{oskarsson2024probabilistic}, without any modifications.

\begin{table}[h!]
\caption{Inference hyperparameters}
\label{tab:inference_hyperparams}
\begin{center}
\begin{tabular}{ll}
\multicolumn{1}{c}{\bf Hyperparameter}  &\multicolumn{1}{c}{\bf Value}
\\ \hline \\
$\sigma_{\text{max}}$       & $80$ \\
$\sigma_{\text{min}}$       & $0.03$ \\
$\rho$                      & $7$ \\
\end{tabular}
\end{center}
\end{table}

\section{Metrics}\label{apx:metrics}
Given a $S$ forecasts we define the RMSE of variable $d$ at step $t$ for the ensemble mean $\bar{\hat{X}}^{s, t}_{g, d}$ at the spatial position $g \in G_{I}$ as
$$
    \text{RMSE}^t_d = \sqrt{\frac{1}{S|G_I|} \sum^S_{s=1} \sum_{g \in G_{I}} (\bar{\hat{X}}^{s, t}_{g, d} - X^{s, t}_{g, d})^2},
$$
where
$$
\bar{\hat{X}}^{s, t}_{n, d} = \frac{1}{N_{\text{ens}}} \sum^{N_{\text{ens}}}_{\text{ens}=1}  \hat{X}^{s, t}_{g, d, \text{ens}},
$$
where $\hat{X}^t_{g, d, \text{ens}}$ is the prediction of ensemble member $\text{ens}$ with a total number of $N_{\text{ens}}$ ensemble members. Note, we follow the standard convention and the WeatherBench 2 benchmark \citet{rasp2023weatherbench} and apply the square root after sample averaging.

To measure the calibration of the uncertainty in the forecasts we use the bias corrected spread-skill ratio for variable $d$ at step $t$ as
$$
\text{SSR}^t_d = \sqrt{\frac{N_{\text{ens}} + 1}{N_{\text{ens}}}}  \frac{\text{Spread}^t_d}{\text{RMSE}^t_d},
$$
where
$$
\text{Spread}^t_d = \sqrt{\frac{1}{S|G_I|N_{\text{ens}} \sum^S_{s=1}\sum_{g \in G_{I}}} \sum^{N_{\text{ens}}}_{\text{ens}=1} (\bar{\hat{X}}^{s, t}_{n, d} - \hat{X}^{s, t}_{g, d, \text{ens}})^2}.
$$
If the uncertainty in the forecasts is well calibrated $\text{SpSkR}^t_d \approx 1$ \citep{fortin2014should}.

We also compute CRPS \citep{gneiting2007strictly} for variable $d$ at step $t$ 
$$
\text{CRPS}^t_d = \frac{1}{S|G_I|N_\text{ens}} \sum^S_{s=1}\sum_{g \in G_{I}} (\sum^{N_{\text{ens}}}_{\text{ens}=1} \lvert \hat{X}^{s, t}_{g, d, \text{ens}} - X^{s, t}_{g, d} \rvert - \frac{1}{2(N_{\text{ens}}-1)}\sum^{N_{\text{ens}}}_{\text{ens}=1}\sum^{N_{\text{ens}}}_{\text{ens}^*=1}\lvert \hat{X}^{s, t}_{g, d, \text{ens}} - \hat{X}^{s, t}_{g, d, \text{ens}^*} \rvert).
$$
Note, we follow the convention of \citet{oskarsson2024probabilistic} and compute the CRPS as a finite sample estimate \citep{zamo2018estimation} over all ensemble members without accounting for any covariance structure.

When calculating metrics for each individual variable separately, we first unnormalize the predictions before comparing them to the ground truth. However, when evaluating the mean performance across all variables, we compute the metrics using normalized data and forecasts. In this case, we normalize the ground truth and compare it to the normalized predictions. The mean normalized score is then obtained by averaging the metric values (RMSE, CRPS, SSR) across all variables.

\section{Additional results}\label{apx:detailed_results}
\begin{figure}[h!]
    \centering
    \includegraphics[width=\textwidth]{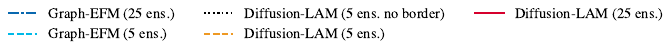}
    \begin{subfigure}[b]{0.3\textwidth}
        \centering
        \includegraphics[width=\textwidth]{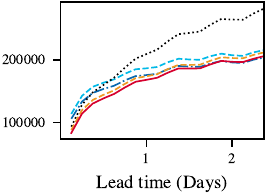}
        \caption{\texttt{nlwrs\_0}}
        % \label{fig:subfigure1}
    \end{subfigure}
    \hfill
    \begin{subfigure}[b]{0.3\textwidth}
        \centering
        \includegraphics[width=\textwidth]{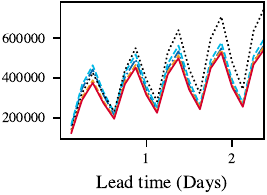}
        \caption{\texttt{nswrs\_0}}
    \end{subfigure}
    \hfill
    \begin{subfigure}[b]{0.3\textwidth}
        \centering
        \includegraphics[width=\textwidth]{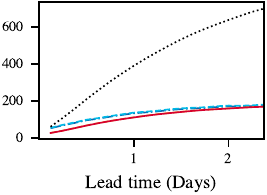}
        \caption{\texttt{pres\_0g}}
    \end{subfigure}
    \hfill
    \begin{subfigure}[b]{0.3\textwidth}
        \centering
        \includegraphics[width=\textwidth]{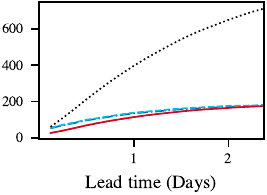}
        \caption{\texttt{pres\_0s}}
    \end{subfigure}
    \hfill
    \begin{subfigure}[b]{0.3\textwidth}
        \centering
        \includegraphics[width=\textwidth]{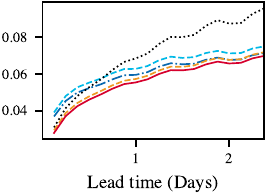}
        \caption{\texttt{r\_2}}
    \end{subfigure}
    \hfill
    \begin{subfigure}[b]{0.3\textwidth}
        \centering
        \includegraphics[width=\textwidth]{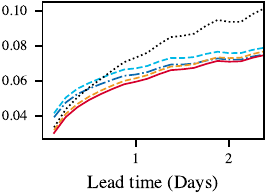}
        \caption{\texttt{r\_65}}
    \end{subfigure}
    \hfill
    \begin{subfigure}[b]{0.3\textwidth}
        \centering
        \includegraphics[width=\textwidth]{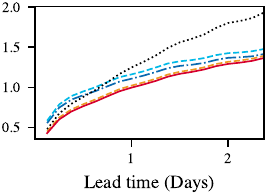}
        \caption{\texttt{t\_2}}
    \end{subfigure}
    \hfill
    \begin{subfigure}[b]{0.3\textwidth}
        \centering
        \includegraphics[width=\textwidth]{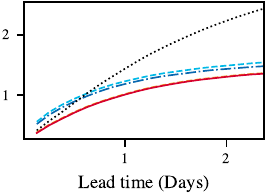}
        \caption{\texttt{t\_500}}
    \end{subfigure}
    \hfill
    \begin{subfigure}[b]{0.3\textwidth}
        \centering
        \includegraphics[width=\textwidth]{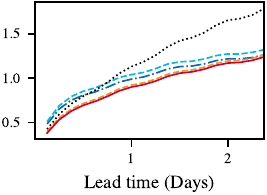}
        \caption{\texttt{t\_65}}
    \end{subfigure}
    \hfill
    \begin{subfigure}[b]{0.3\textwidth}
        \centering
        \includegraphics[width=\textwidth]{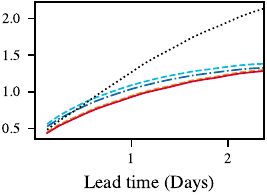}
        \caption{\texttt{t\_850}}
    \end{subfigure}
    \hfill
    \begin{subfigure}[b]{0.3\textwidth}
        \centering
        \includegraphics[width=\textwidth]{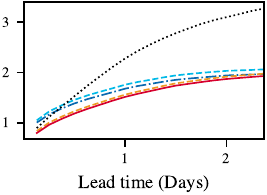}
        \caption{\texttt{u\_65}}
    \end{subfigure}
    \hfill
    \begin{subfigure}[b]{0.3\textwidth}
        \centering
        \includegraphics[width=\textwidth]{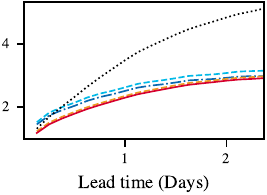}
        \caption{\texttt{u\_850}}
    \end{subfigure}
    \hfill
    \begin{subfigure}[b]{0.3\textwidth}
        \centering
        \includegraphics[width=\textwidth]{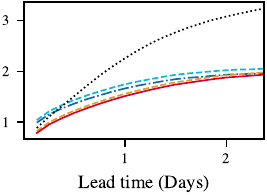}
        \caption{\texttt{v\_65}}
    \end{subfigure}
    \hfill
    \begin{subfigure}[b]{0.3\textwidth}
        \centering
        \includegraphics[width=\textwidth]{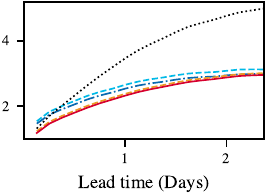}
        \caption{\texttt{v\_850}}
    \end{subfigure}
    \hfill
    \begin{subfigure}[b]{0.3\textwidth}
        \centering
        \includegraphics[width=\textwidth]{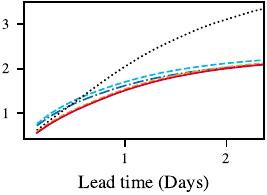}
        \caption{\texttt{wvint\_0}}
    \end{subfigure}
    \hfill
    \begin{subfigure}[b]{0.3\textwidth}
        \centering
        \includegraphics[width=\textwidth]{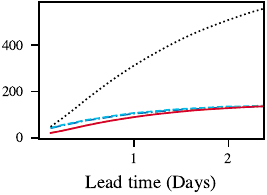}
        \caption{\texttt{z\_1000}}
    \end{subfigure}
    \begin{subfigure}[b]{0.3\textwidth}
        \centering
        \includegraphics[width=\textwidth]{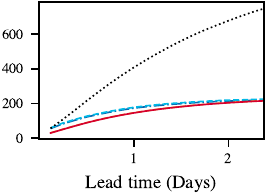}
        \caption{\texttt{z\_500}}
    \end{subfigure}
    \hfill
    \caption{The RMSE results for each variable.}
    \label{fig:rmse_all}
\end{figure}

\begin{figure}[h!]
    \centering
    \includegraphics[width=\textwidth]{figures/crps/crps_legend.pdf}
    \begin{subfigure}[b]{0.3\textwidth}
        \centering
        \includegraphics[width=\textwidth]{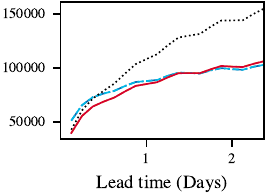}
        \caption{\texttt{nlwrs\_0}}
        % \label{fig:subfigure1}
    \end{subfigure}
    \hfill
    \begin{subfigure}[b]{0.3\textwidth}
        \centering
        \includegraphics[width=\textwidth]{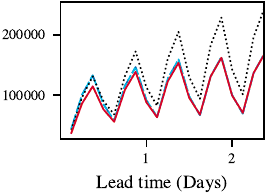}
        \caption{\texttt{nswrs\_0}}
    \end{subfigure}
    \hfill
    \begin{subfigure}[b]{0.3\textwidth}
        \centering
        \includegraphics[width=\textwidth]{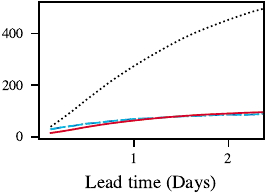}
        \caption{\texttt{pres\_0g}}
    \end{subfigure}
    \hfill
    \begin{subfigure}[b]{0.3\textwidth}
        \centering
        \includegraphics[width=\textwidth]{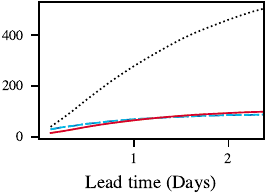}
        \caption{\texttt{pres\_0s}}
    \end{subfigure}
    \hfill
    \begin{subfigure}[b]{0.3\textwidth}
        \centering
        \includegraphics[width=\textwidth]{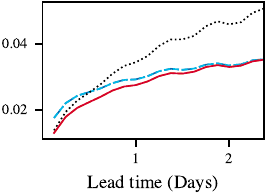}
        \caption{\texttt{r\_2}}
    \end{subfigure}
    \hfill
    \begin{subfigure}[b]{0.3\textwidth}
        \centering
        \includegraphics[width=\textwidth]{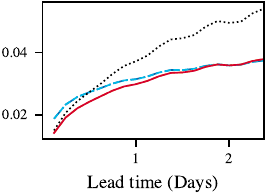}
        \caption{\texttt{r\_65}}
    \end{subfigure}
    \hfill
    \begin{subfigure}[b]{0.3\textwidth}
        \centering
        \includegraphics[width=\textwidth]{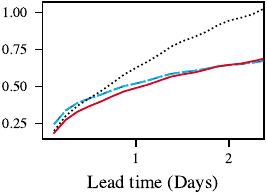}
        \caption{\texttt{t\_2}}
    \end{subfigure}
    \hfill
    \begin{subfigure}[b]{0.3\textwidth}
        \centering
        \includegraphics[width=\textwidth]{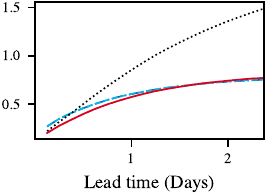}
        \caption{\texttt{t\_500}}
    \end{subfigure}
    \hfill
    \begin{subfigure}[b]{0.3\textwidth}
        \centering
        \includegraphics[width=\textwidth]{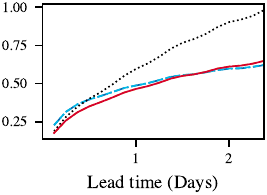}
        \caption{\texttt{t\_65}}
    \end{subfigure}
    \hfill
    \begin{subfigure}[b]{0.3\textwidth}
        \centering
        \includegraphics[width=\textwidth]{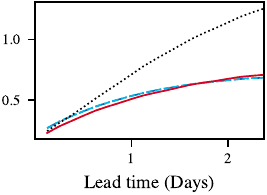}
        \caption{\texttt{t\_850}}
    \end{subfigure}
    \hfill
    \begin{subfigure}[b]{0.3\textwidth}
        \centering
        \includegraphics[width=\textwidth]{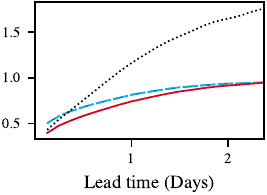}
        \caption{\texttt{u\_65}}
    \end{subfigure}
    \hfill
    \begin{subfigure}[b]{0.3\textwidth}
        \centering
        \includegraphics[width=\textwidth]{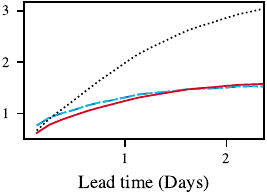}
        \caption{\texttt{u\_850}}
    \end{subfigure}
    \hfill
    \begin{subfigure}[b]{0.3\textwidth}
        \centering
        \includegraphics[width=\textwidth]{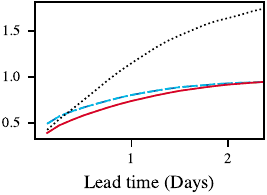}
        \caption{\texttt{v\_65}}
    \end{subfigure}
    \hfill
    \begin{subfigure}[b]{0.3\textwidth}
        \centering
        \includegraphics[width=\textwidth]{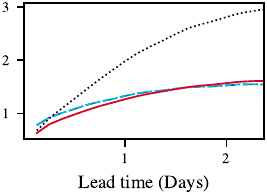}
        \caption{\texttt{v\_850}}
    \end{subfigure}
    \hfill
    \begin{subfigure}[b]{0.3\textwidth}
        \centering
        \includegraphics[width=\textwidth]{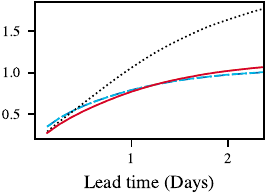}
        \caption{\texttt{wvint\_0}}
    \end{subfigure}
    \hfill
    \begin{subfigure}[b]{0.3\textwidth}
        \centering
        \includegraphics[width=\textwidth]{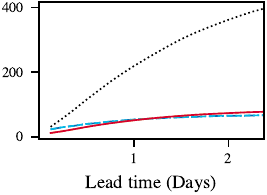}
        \caption{\texttt{z\_1000}}
    \end{subfigure}
    \begin{subfigure}[b]{0.3\textwidth}
        \centering
        \includegraphics[width=\textwidth]{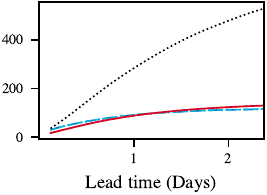}
        \caption{\texttt{z\_500}}
    \end{subfigure}
    \hfill
    \caption{The CRPS results for each variable.}
    \label{fig:crps_all}
\end{figure}

\begin{figure}[h!]
    \centering
    \includegraphics[width=\textwidth]{figures/spskr/spskr_legend.pdf}
    \begin{subfigure}[b]{0.3\textwidth}
        \centering
        \includegraphics[width=\textwidth]{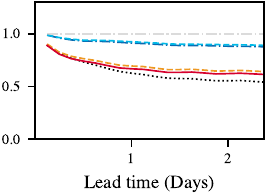}
        \caption{\texttt{nlwrs\_0}}
    \end{subfigure}
    \hfill
    \begin{subfigure}[b]{0.3\textwidth}
        \centering
        \includegraphics[width=\textwidth]{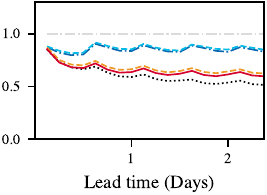}
        \caption{\texttt{nswrs\_0}}
    \end{subfigure}
    \hfill
    \begin{subfigure}[b]{0.3\textwidth}
        \centering
        \includegraphics[width=\textwidth]{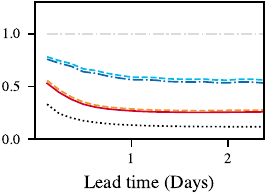}
        \caption{\texttt{pres\_0g}}
    \end{subfigure}
    \hfill
    \begin{subfigure}[b]{0.3\textwidth}
        \centering
        \includegraphics[width=\textwidth]{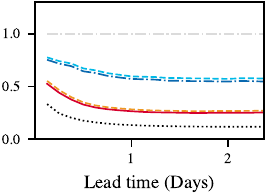}
        \caption{\texttt{pres\_0s}}
    \end{subfigure}
    \hfill
    \begin{subfigure}[b]{0.3\textwidth}
        \centering
        \includegraphics[width=\textwidth]{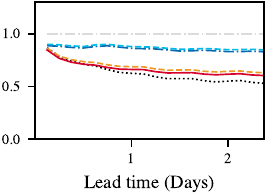}
        \caption{\texttt{r\_2}}
    \end{subfigure}
    \hfill
    \begin{subfigure}[b]{0.3\textwidth}
        \centering
        \includegraphics[width=\textwidth]{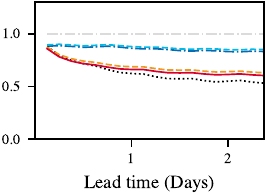}
        \caption{\texttt{r\_65}}
    \end{subfigure}
    \hfill
    \begin{subfigure}[b]{0.3\textwidth}
        \centering
        \includegraphics[width=\textwidth]{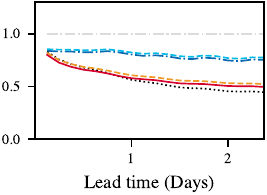}
        \caption{\texttt{t\_2}}
    \end{subfigure}
    \hfill
    \begin{subfigure}[b]{0.3\textwidth}
        \centering
        \includegraphics[width=\textwidth]{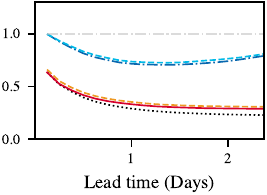}
        \caption{\texttt{t\_500}}
    \end{subfigure}
    \hfill
    \begin{subfigure}[b]{0.3\textwidth}
        \centering
        \includegraphics[width=\textwidth]{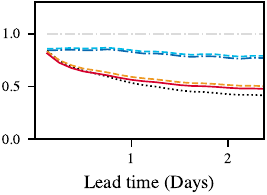}
        \caption{\texttt{t\_65}}
    \end{subfigure}
    \hfill
    \begin{subfigure}[b]{0.3\textwidth}
        \centering
        \includegraphics[width=\textwidth]{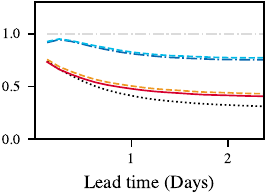}
        \caption{\texttt{t\_850}}
    \end{subfigure}
    \hfill
    \begin{subfigure}[b]{0.3\textwidth}
        \centering
        \includegraphics[width=\textwidth]{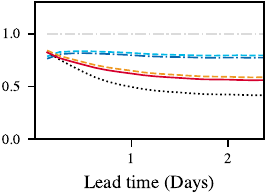}
        \caption{\texttt{u\_65}}
    \end{subfigure}
    \hfill
    \begin{subfigure}[b]{0.3\textwidth}
        \centering
        \includegraphics[width=\textwidth]{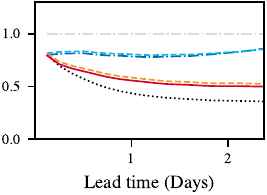}
        \caption{\texttt{u\_850}}
    \end{subfigure}
    \hfill
    \begin{subfigure}[b]{0.3\textwidth}
        \centering
        \includegraphics[width=\textwidth]{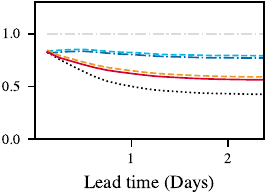}
        \caption{\texttt{v\_65}}
    \end{subfigure}
    \hfill
    \begin{subfigure}[b]{0.3\textwidth}
        \centering
        \includegraphics[width=\textwidth]{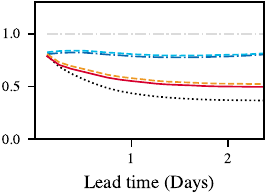}
        \caption{\texttt{v\_850}}
    \end{subfigure}
    \hfill
    \begin{subfigure}[b]{0.3\textwidth}
        \centering
        \includegraphics[width=\textwidth]{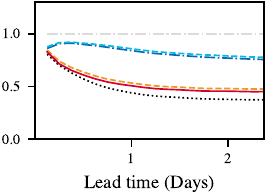}
        \caption{\texttt{wvint\_0}}
    \end{subfigure}
    \hfill
    \begin{subfigure}[b]{0.3\textwidth}
        \centering
        \includegraphics[width=\textwidth]{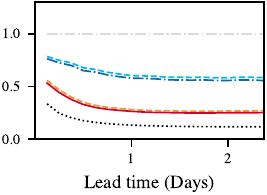}
        \caption{\texttt{z\_1000}}
    \end{subfigure}
    \begin{subfigure}[b]{0.3\textwidth}
        \centering
        \includegraphics[width=\textwidth]{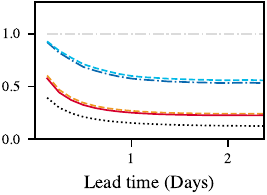}
        \caption{\texttt{z\_500}}
    \end{subfigure}
    \hfill
    \caption{The SSR results for each variable.}
    \label{fig:spskr_all}
\end{figure}

Here we present the detailed results for each variable in \cref{fig:rmse_all}, \cref{fig:crps_all}, \cref{fig:spskr_all} along with a \SI{57}{\hour} forecast for a randomly selected sample from the test set in \cref{fig:samples_all}. For evaluations of deterministic models and less competitive probabilistic baselines on the MEPS dataset, we refer the reader to \citet{oskarsson2023graph-lam, oskarsson2024probabilistic}.

\begin{figure}[h]
    \centering
    \begin{subfigure}[b]{\textwidth}%
        \centering
        \includegraphics[width=\textwidth]{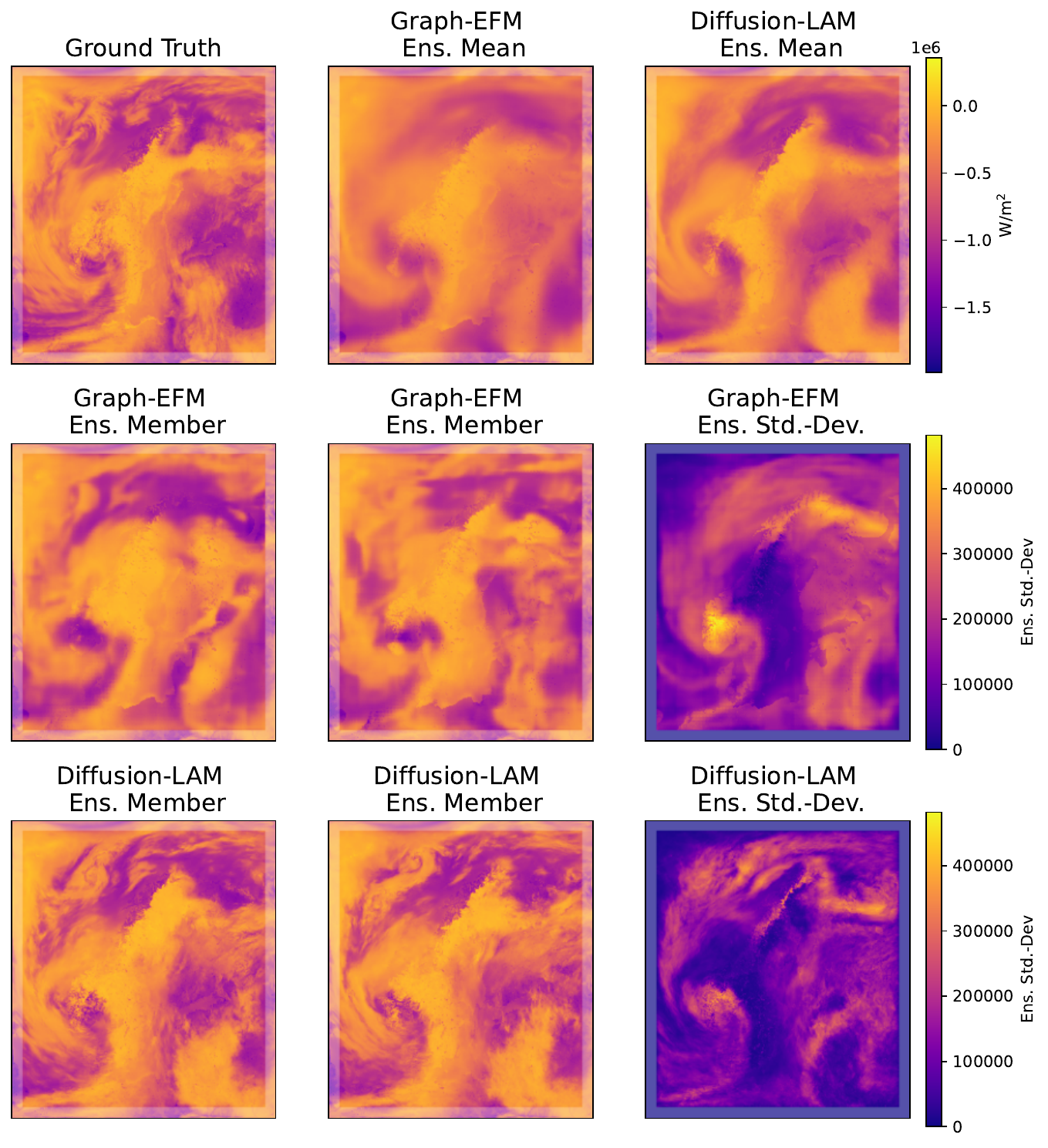}
        \caption{\texttt{nlwrs\_0}}
    \end{subfigure}
\end{figure}
\begin{figure}[tbp]\ContinuedFloat%
    \centering
    \begin{subfigure}[b]{\textwidth}
        \centering
        \includegraphics[width=\textwidth]{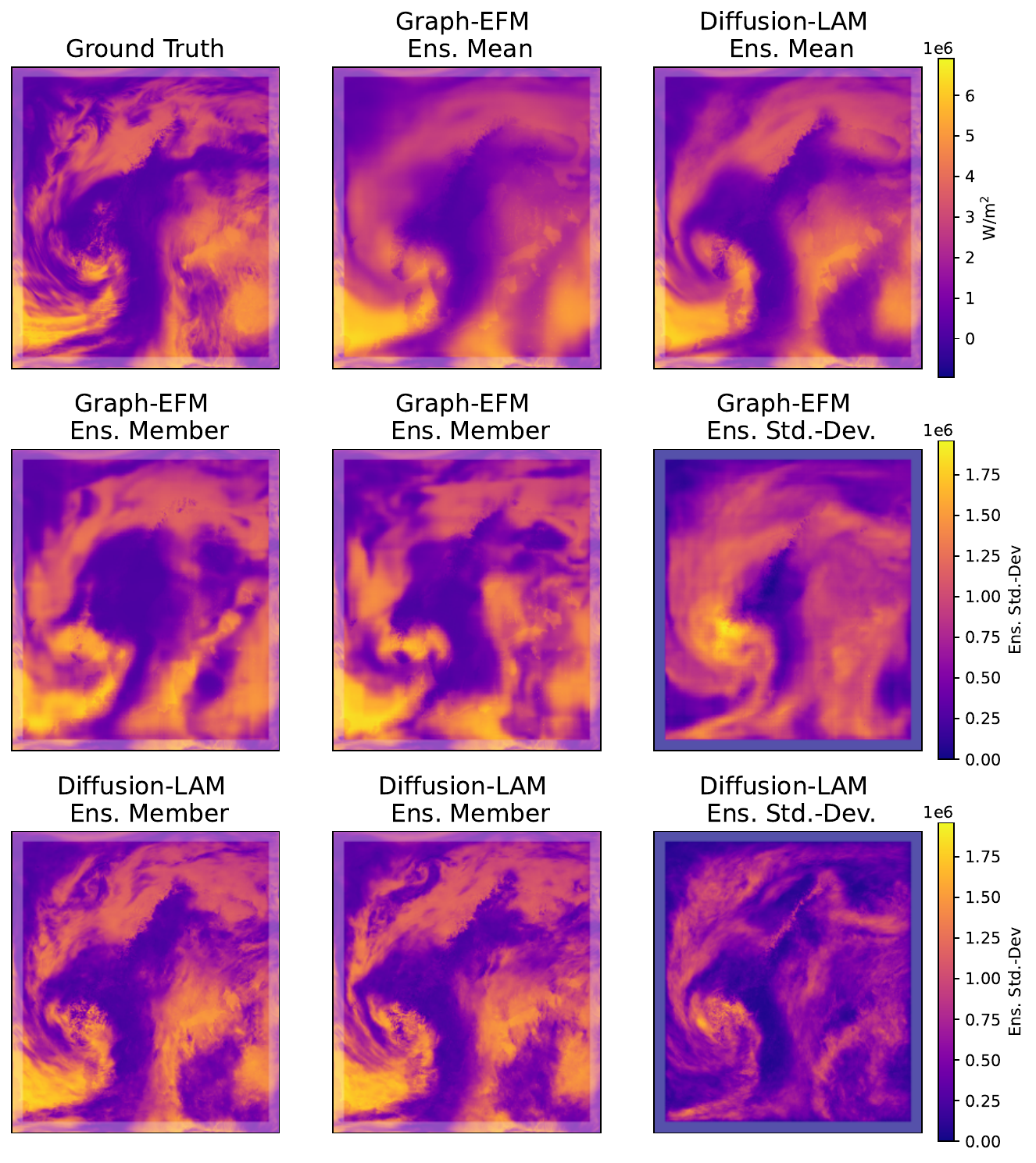}
        \caption{\texttt{nswrs\_0}}
    \end{subfigure}
\end{figure}
\begin{figure}[tbp]\ContinuedFloat
    \centering
    \begin{subfigure}[b]{\textwidth}
        \centering
        \includegraphics[width=\textwidth]{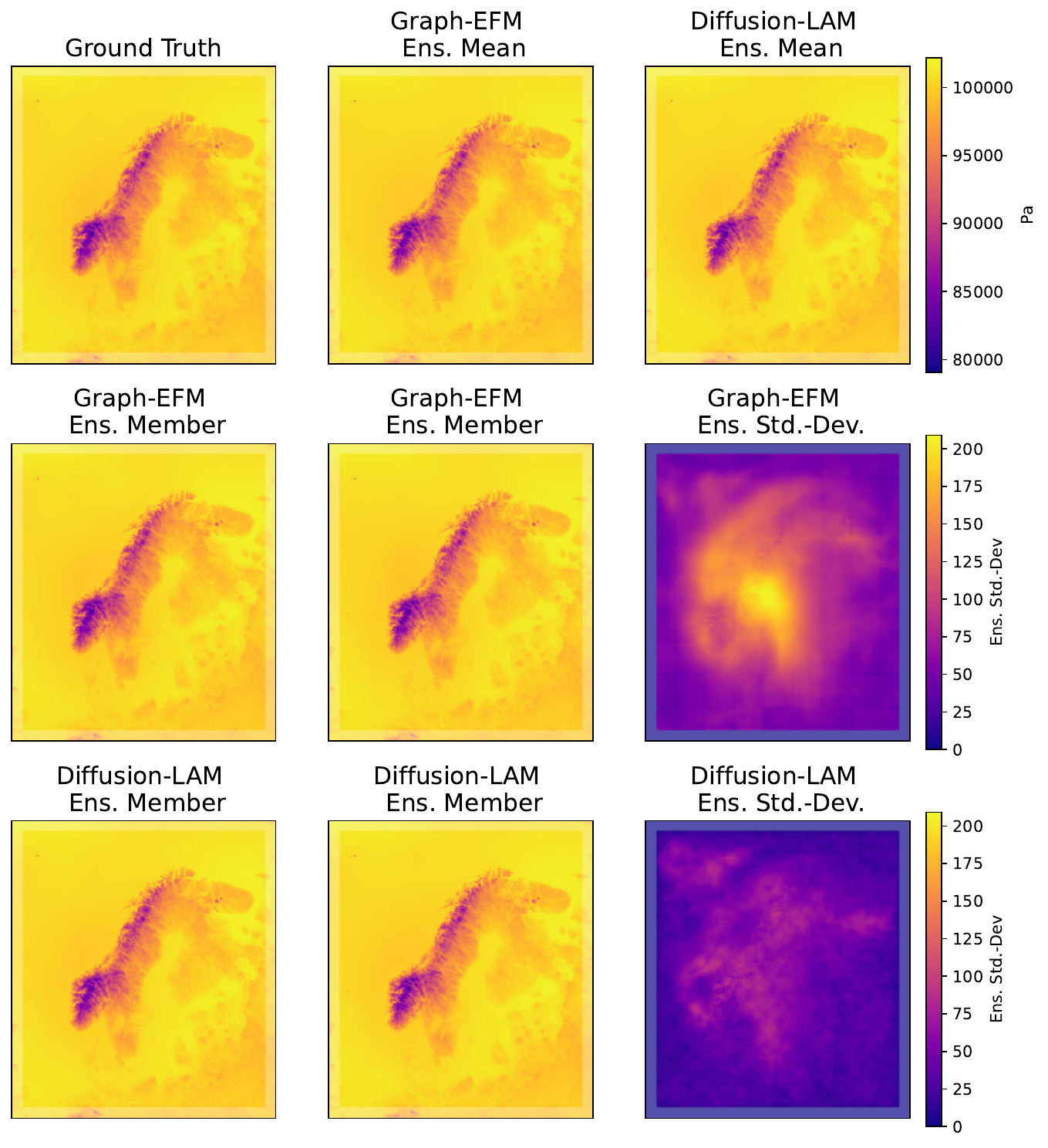}
        \caption{\texttt{pres\_0g}}
    \end{subfigure}
\end{figure}
\begin{figure}[tbp]\ContinuedFloat
    \centering
    \begin{subfigure}[b]{\textwidth}
        \centering
        \includegraphics[width=\textwidth]{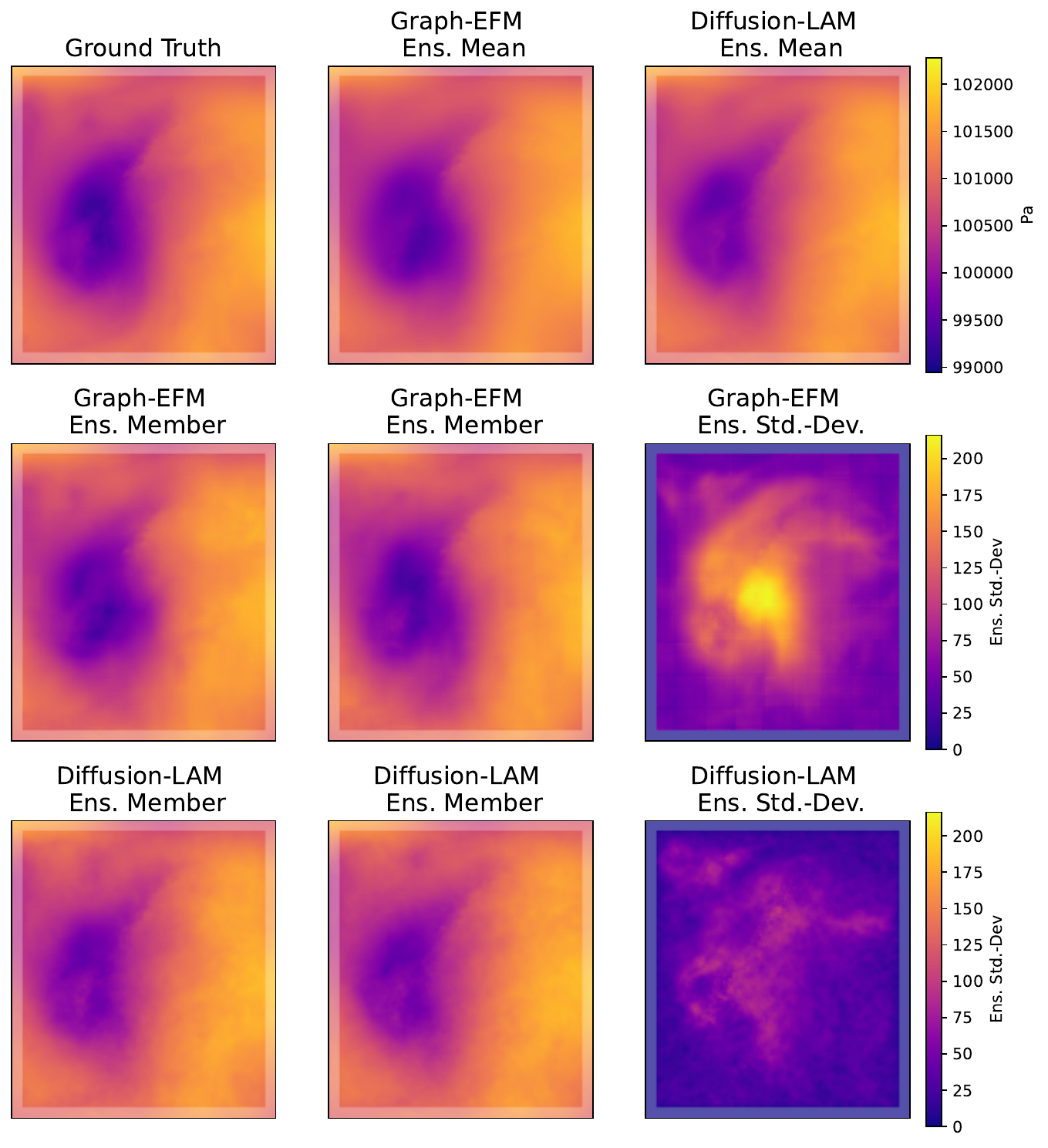}
        \caption{\texttt{pres\_0s}}
    \end{subfigure}
\end{figure}
\begin{figure}[tbp]\ContinuedFloat
    \centering
    \begin{subfigure}[b]{\textwidth}
        \centering
        \includegraphics[width=\textwidth]{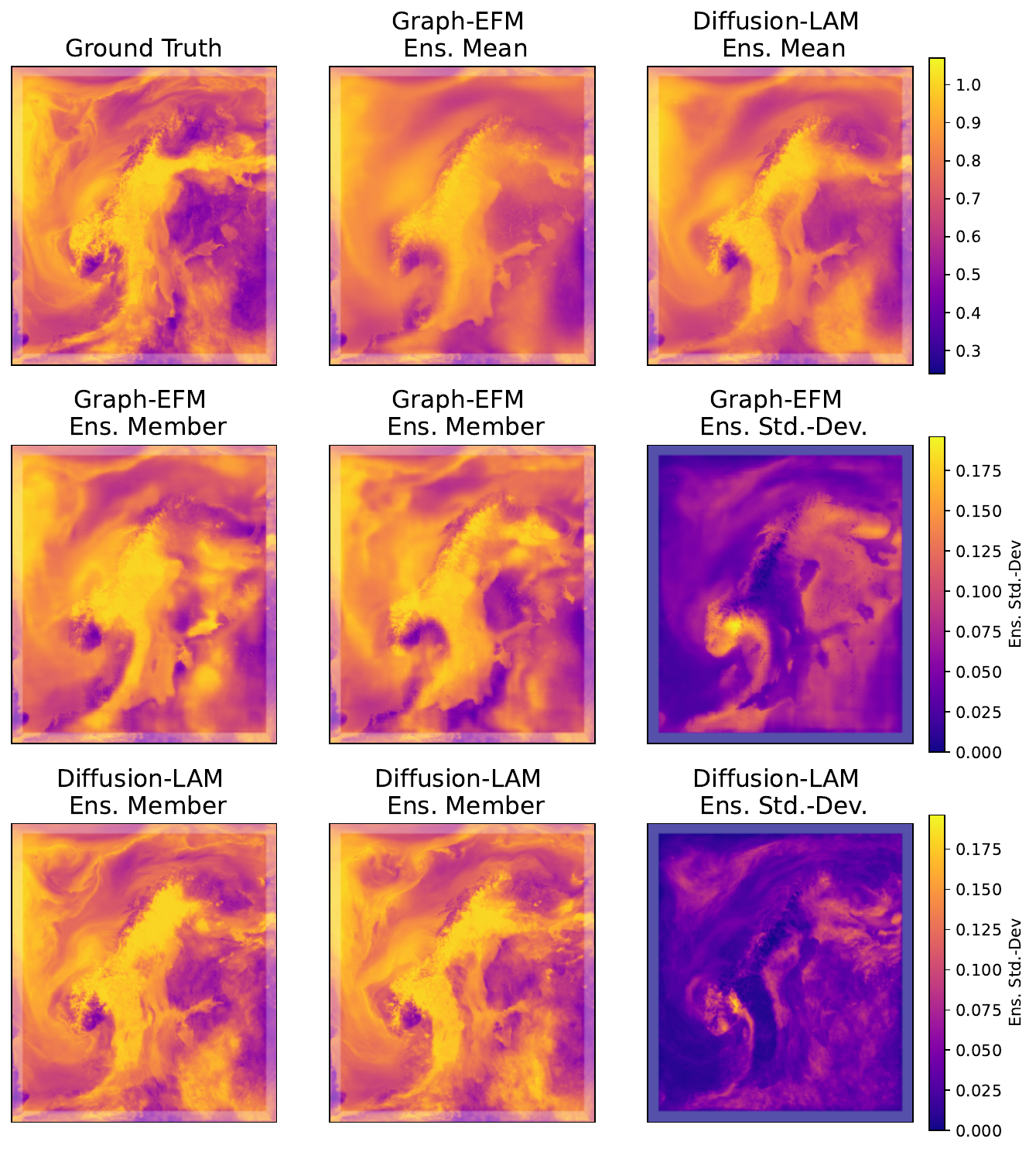}
        \caption{\texttt{r\_2}}
    \end{subfigure}
\end{figure}
\begin{figure}[tbp]\ContinuedFloat
    \centering
    \begin{subfigure}[b]{\textwidth}
        \centering
        \includegraphics[width=\textwidth]{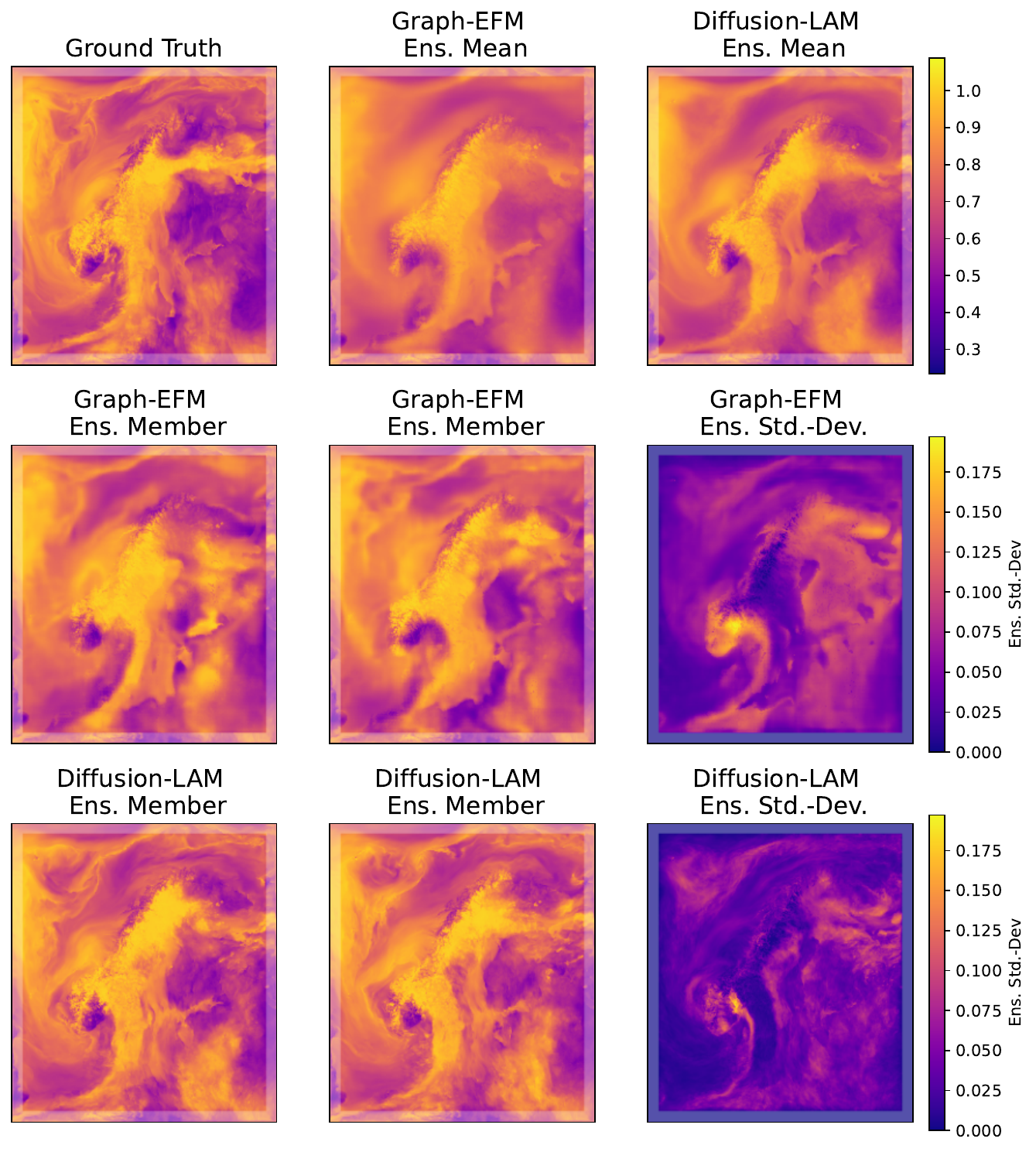}
        \caption{\texttt{r\_65}}
    \end{subfigure}
\end{figure}
\begin{figure}[tbp]\ContinuedFloat
    \centering
    \begin{subfigure}[b]{\textwidth}
        \centering
        \includegraphics[width=\textwidth]{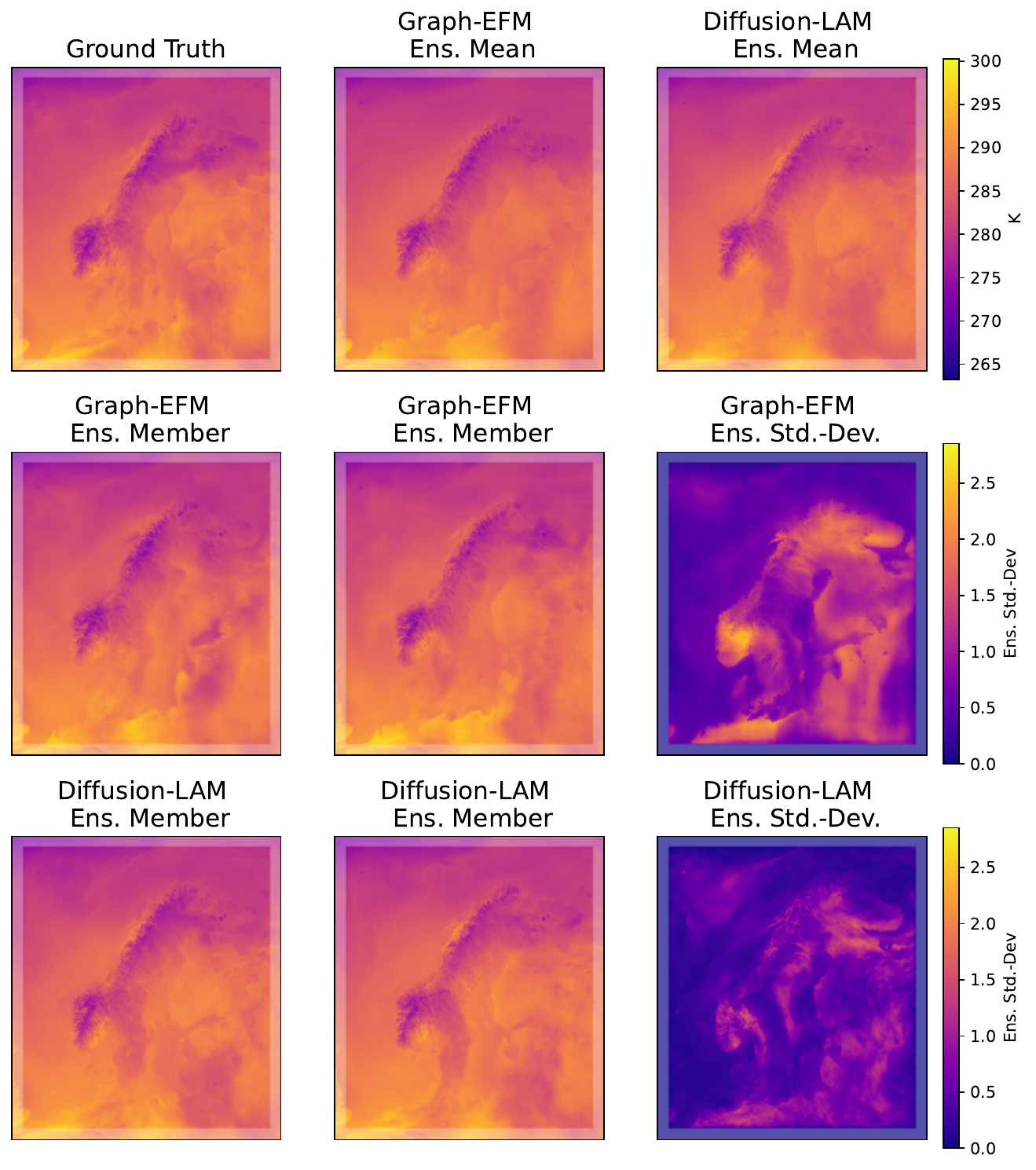}
        \caption{\texttt{t\_2}}
    \end{subfigure}
\end{figure}
\begin{figure}[tbp]\ContinuedFloat
    \centering
    \begin{subfigure}[b]{\textwidth}
        \centering
        \includegraphics[width=\textwidth]{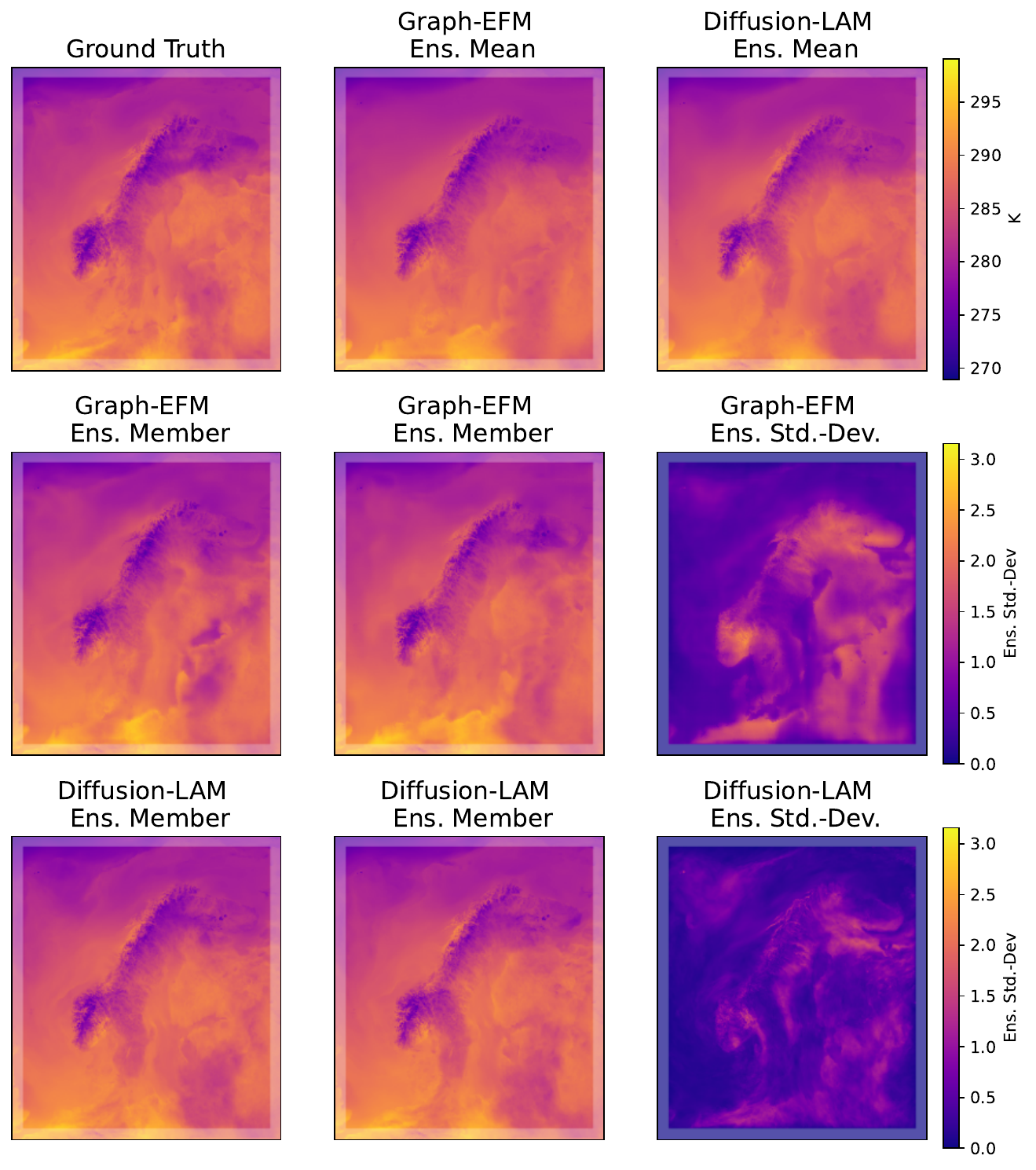}
        \caption{\texttt{t\_500}}
    \end{subfigure}
\end{figure}
\begin{figure}[tbp]\ContinuedFloat
    \centering
    \begin{subfigure}[b]{\textwidth}
        \centering
        \includegraphics[width=\textwidth]{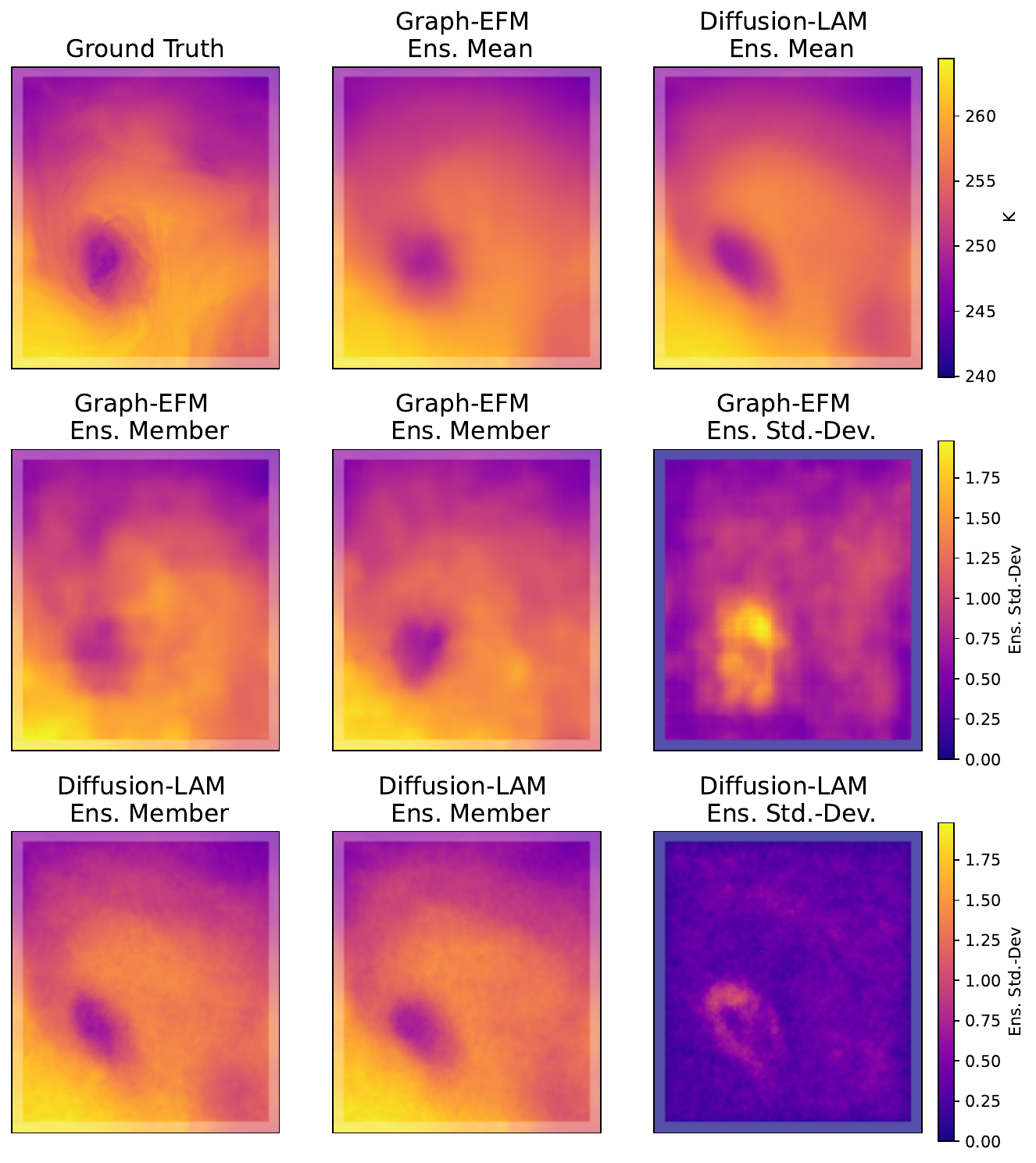}
        \caption{\texttt{t\_65}}
    \end{subfigure}
\end{figure}
\begin{figure}[tbp]\ContinuedFloat
    \centering
    \begin{subfigure}[b]{\textwidth}
        \centering
        \includegraphics[width=\textwidth]{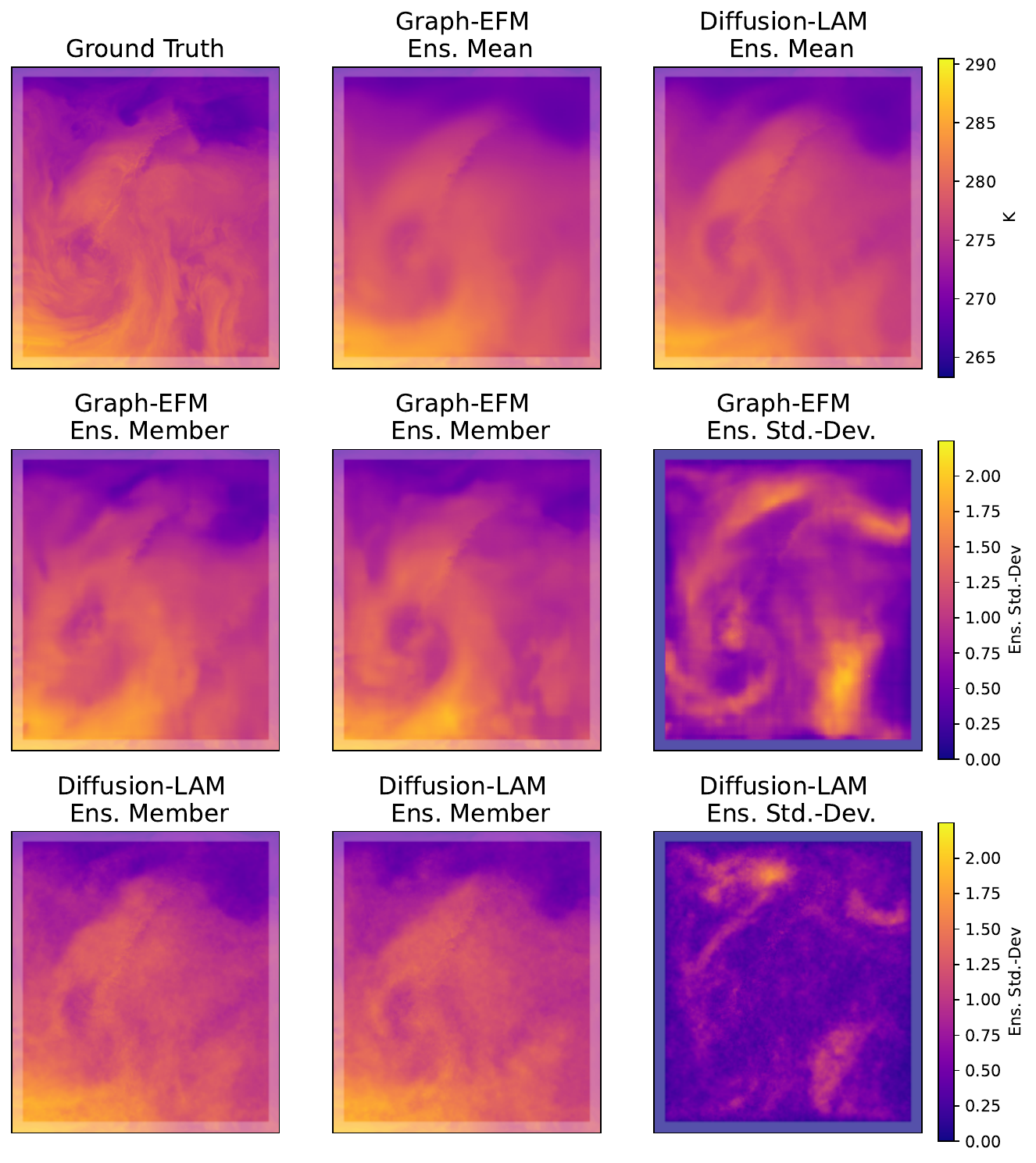}
        \caption{\texttt{t\_850}}
    \end{subfigure}
\end{figure}
\begin{figure}[tbp]\ContinuedFloat
    \centering
    \begin{subfigure}[b]{\textwidth}
        \centering
        \includegraphics[width=\textwidth]{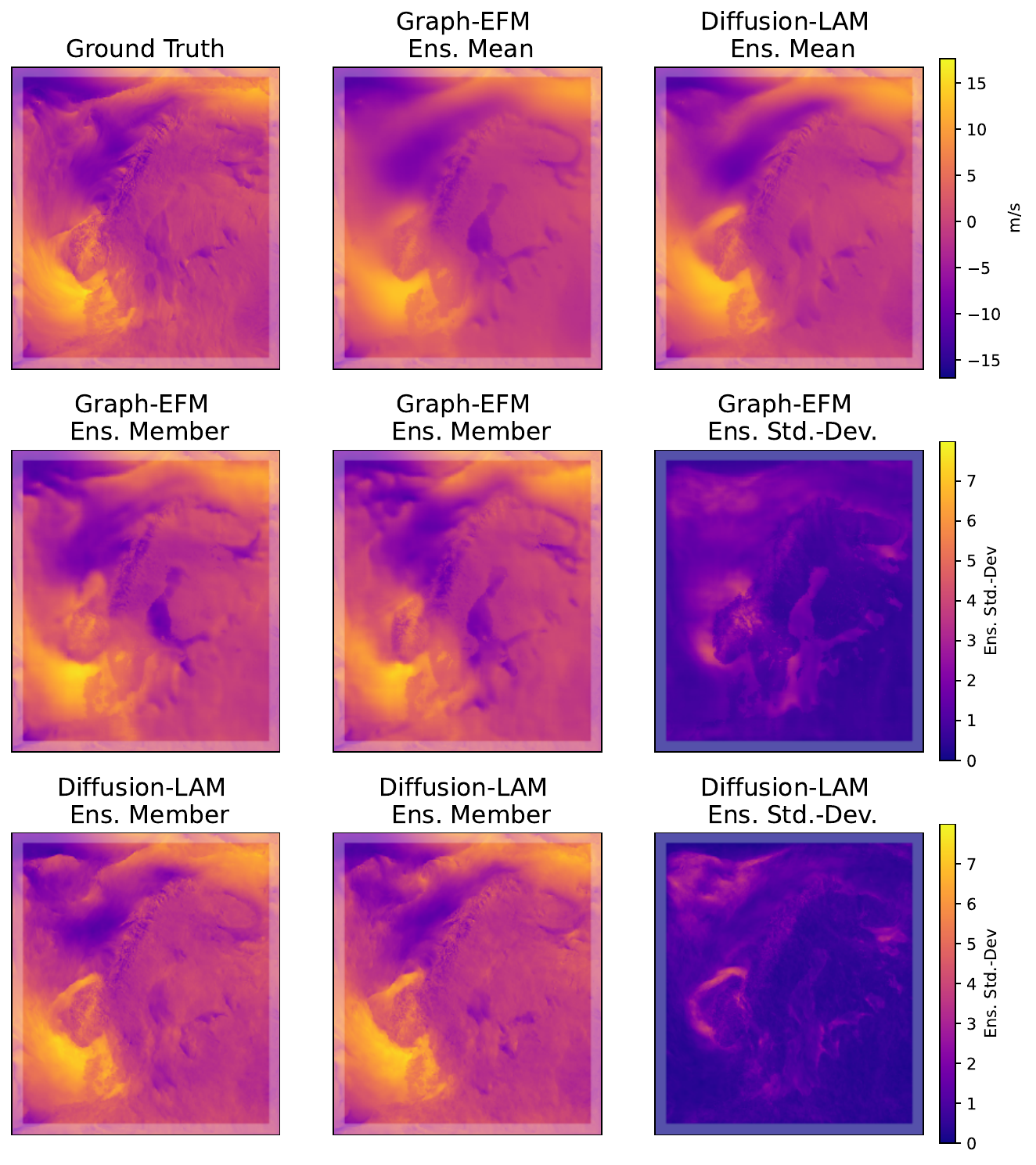}
        \caption{\texttt{u\_65}}
    \end{subfigure}
\end{figure}
\begin{figure}[tbp]\ContinuedFloat
    \centering
    \begin{subfigure}[b]{\textwidth}
        \centering
        \includegraphics[width=\textwidth]{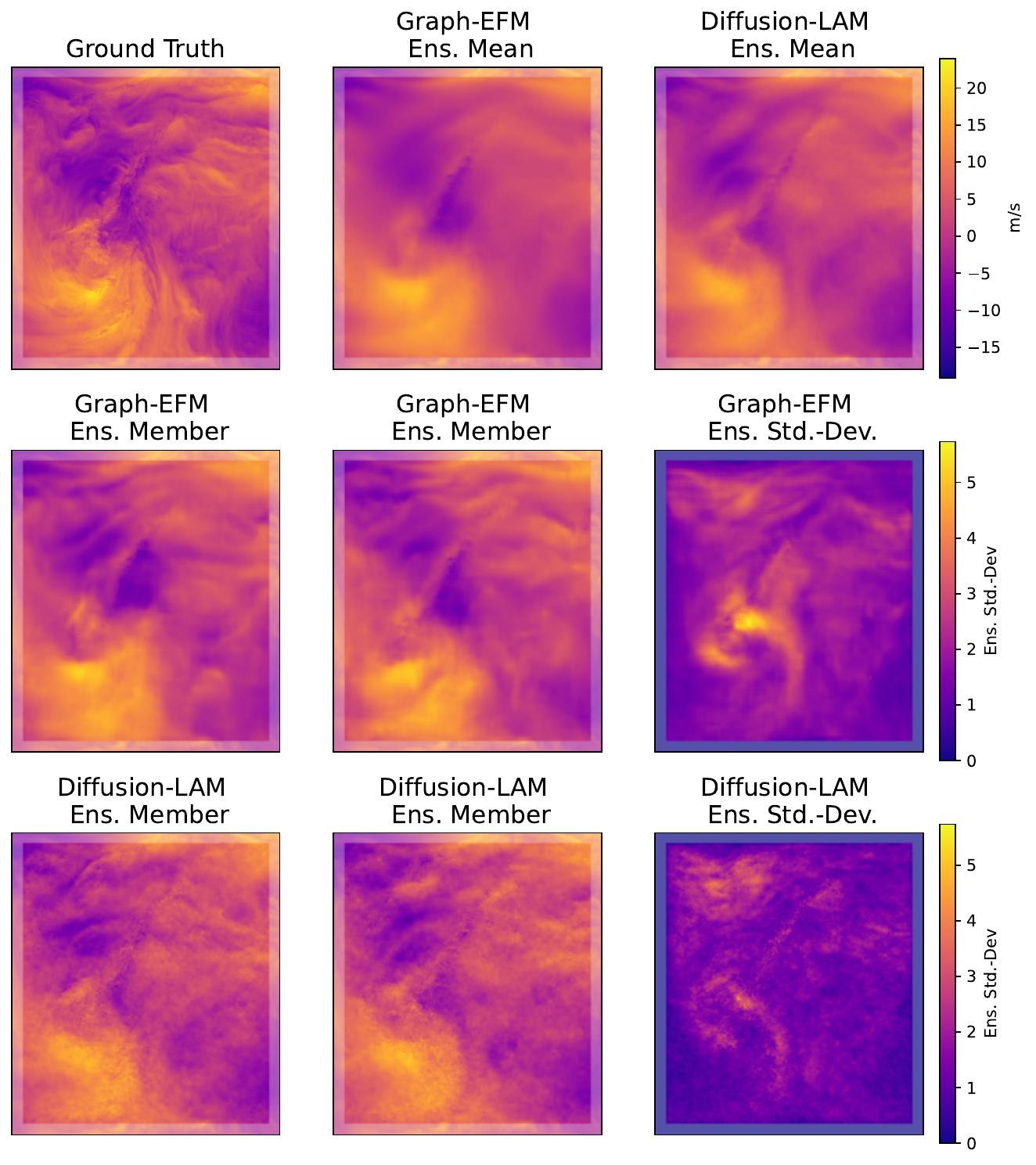}
        \caption{\texttt{u\_850}}
    \end{subfigure}
\end{figure}
\begin{figure}[tbp]\ContinuedFloat
    \centering
    \begin{subfigure}[b]{\textwidth}
        \centering
        \includegraphics[width=\textwidth]{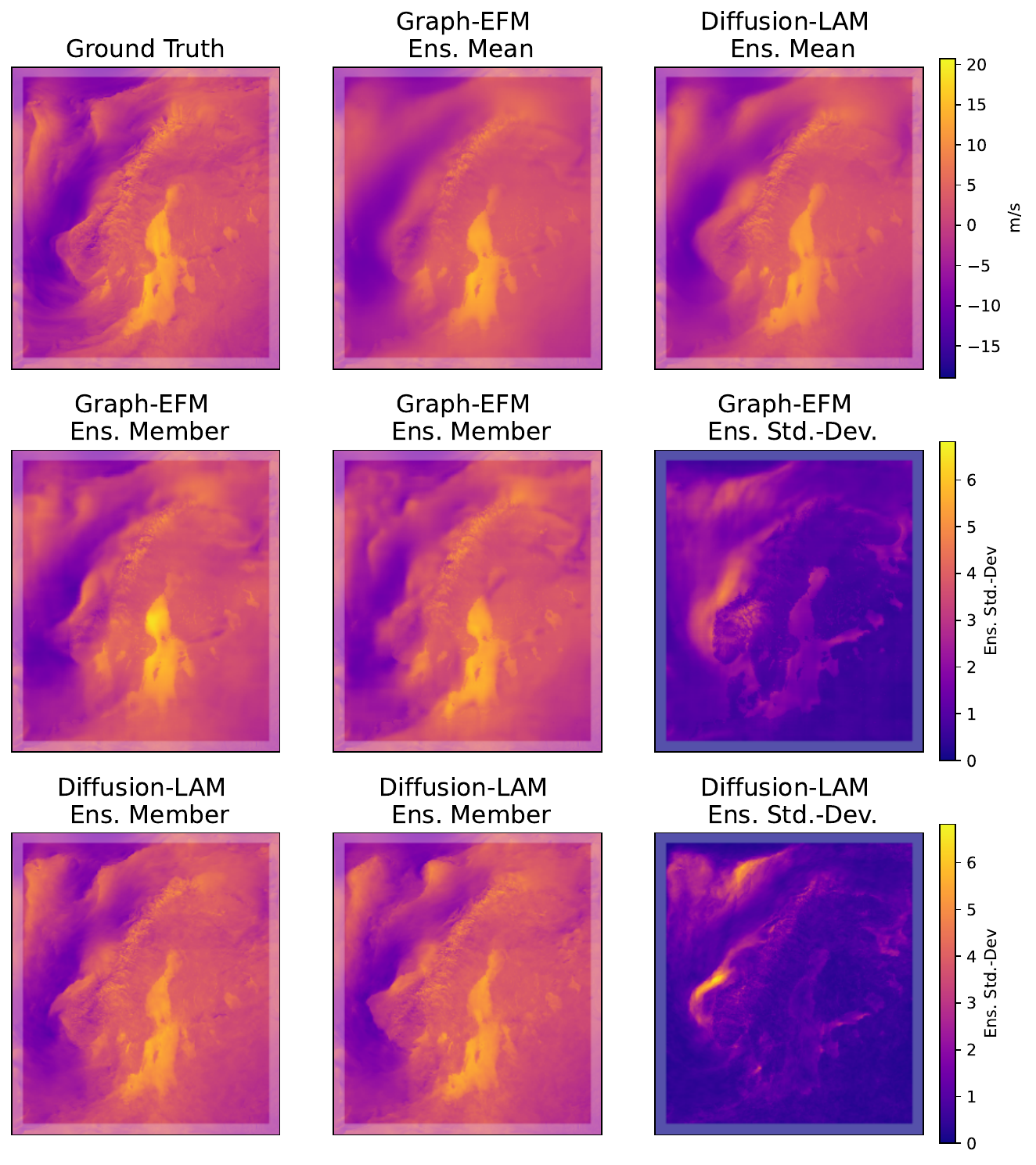}
        \caption{\texttt{v\_65}}
    \end{subfigure}
\end{figure}
\begin{figure}[tbp]\ContinuedFloat
    \centering
    \begin{subfigure}[b]{\textwidth}
        \centering
        \includegraphics[width=\textwidth]{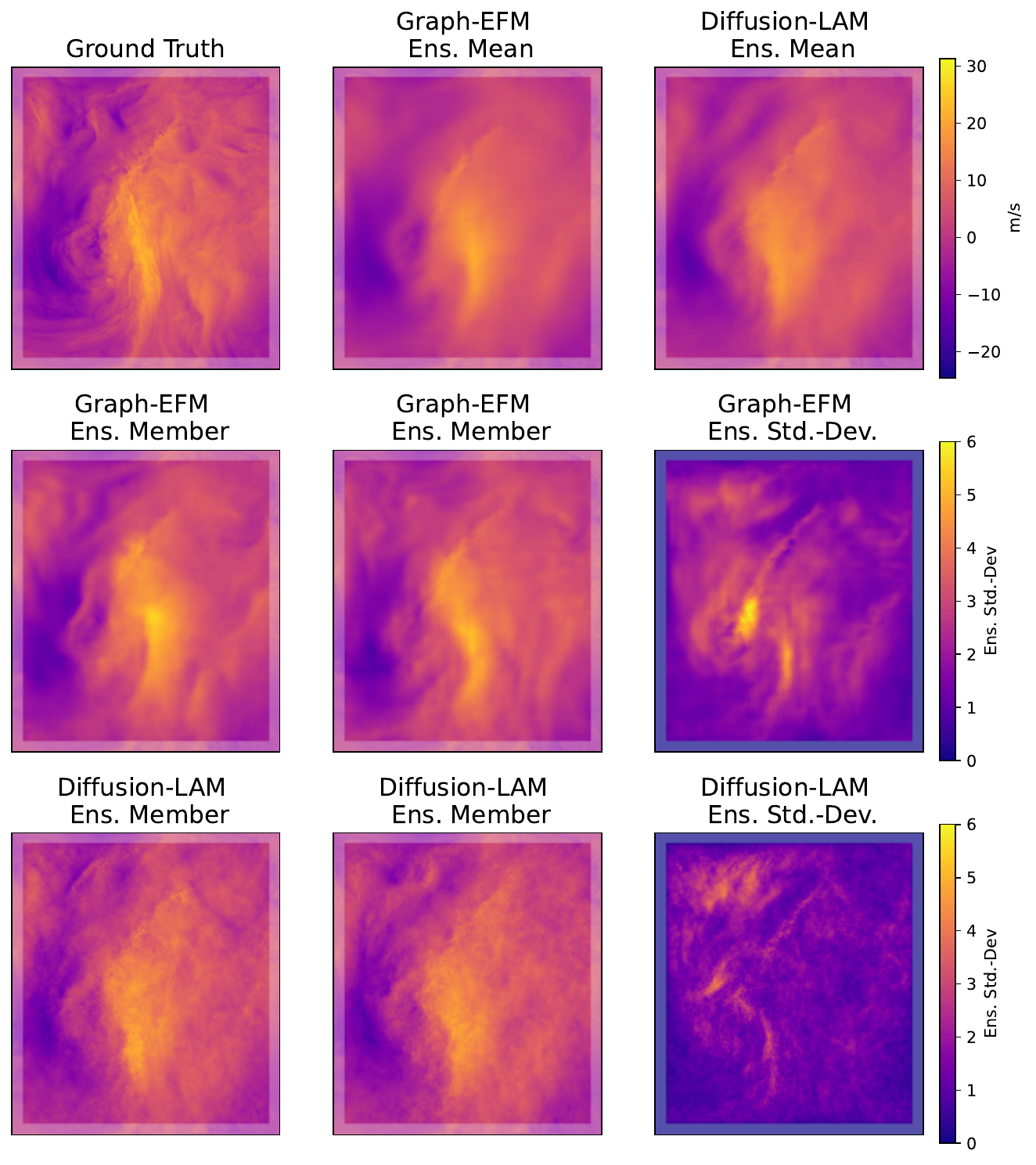}
        \caption{\texttt{v\_850}}
    \end{subfigure}
\end{figure}
\begin{figure}[tbp]\ContinuedFloat
    \centering
    \begin{subfigure}[b]{\textwidth}
        \centering
        \includegraphics[width=\textwidth]{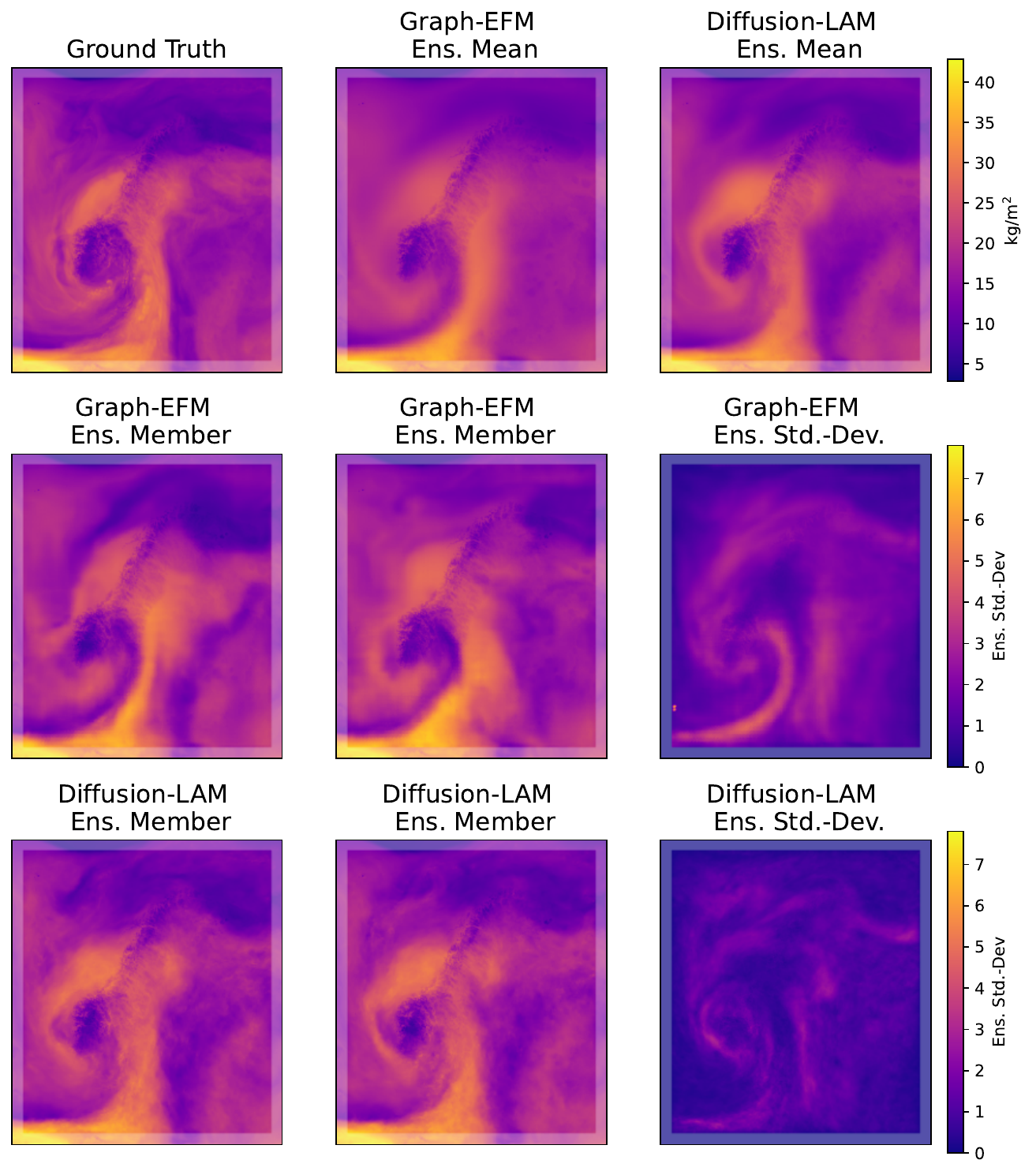}
        \caption{\texttt{wvint\_0}}
    \end{subfigure}
\end{figure}
\begin{figure}[tbp]\ContinuedFloat
    \centering
    \begin{subfigure}[b]{\textwidth}
        \centering
        \includegraphics[width=\textwidth]{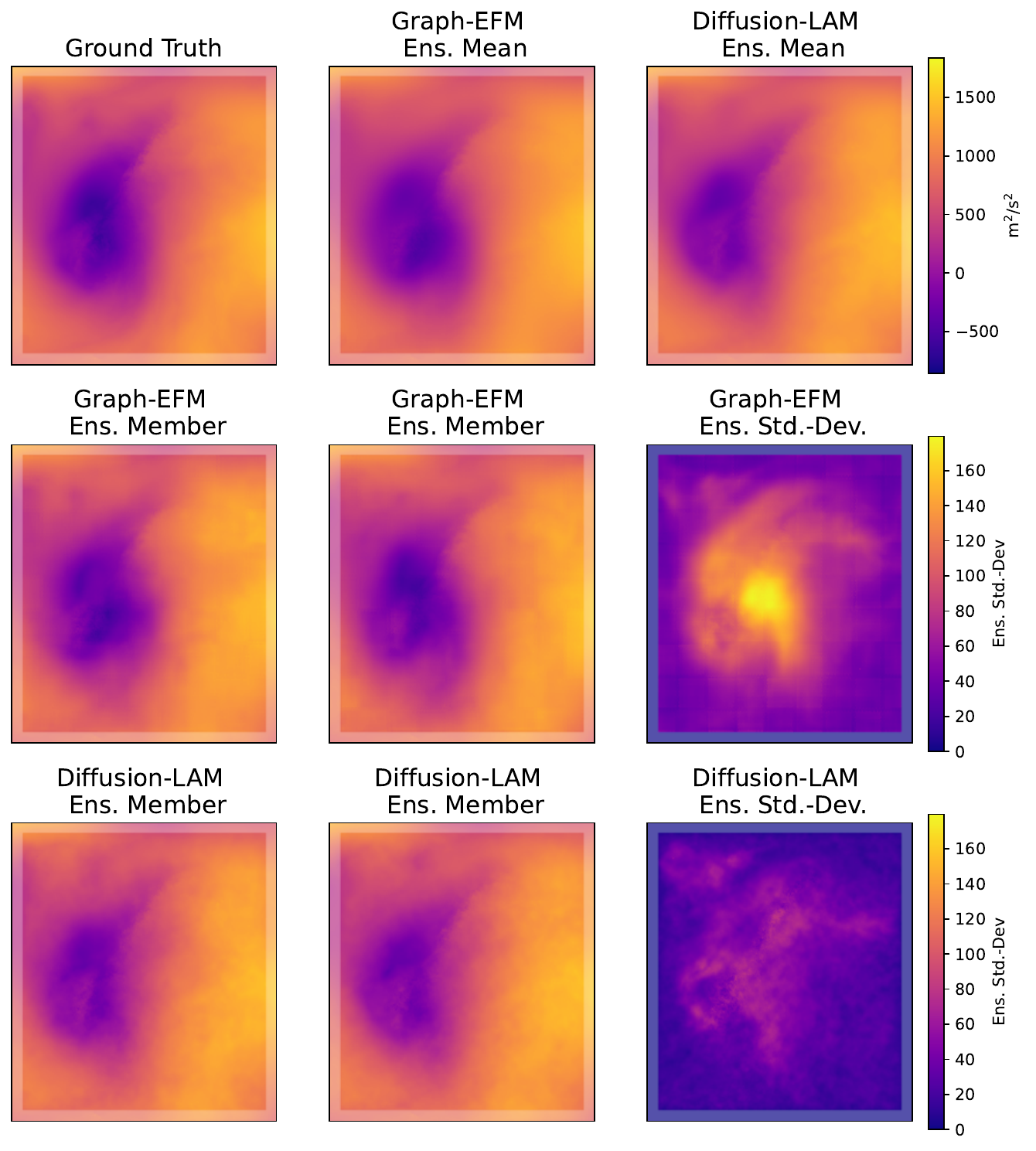}
        \caption{\texttt{z\_1000}}
    \end{subfigure}
\end{figure}
\begin{figure}[tbp]\ContinuedFloat
    \centering
    \begin{subfigure}[b]{\textwidth}
        \centering
        \includegraphics[width=\textwidth]{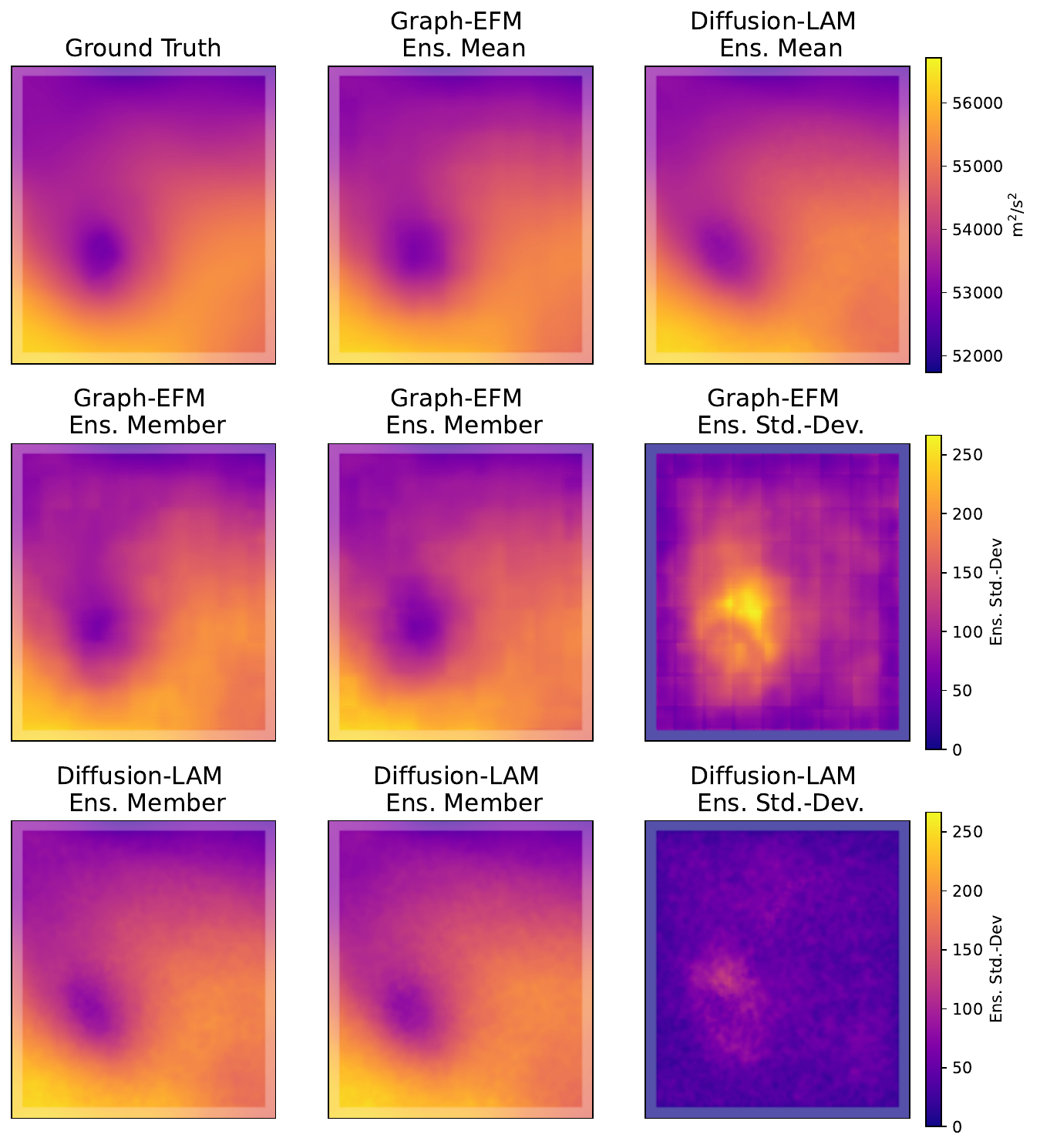}
        \caption{\texttt{z\_500}}
    \end{subfigure}
    \caption{An ensemble forecasts with Diffusion-LAM for each variable at \SI{57}{\hour}.}
    \label{fig:samples_all}
\end{figure}

\section{Future work}\label{apx:future_work}
In this work, we aim to develop an emulator model for the MEPS forecasting system. However, several promising directions for future research remain. One interesting direction would be to design a model initialized directly from analysis or observational data, enabling direct comparison with reanalysis results. Additionally, incorporating boundary information from a global model, potentially using different resolutions, variables, or timeframes could be interesting LAM research. Developing LAMs with higher spatial and temporal resolution is also a worthwhile pursuit to better capture the underlying physical dynamics and to make the forecasts more valuable.  

Further improvements could target the diffusion model’s sampling efficiency. Exploring faster sampling methods, such as consistency models, or strategies to increase the SSR without oversmoothing ensemble members or introducing non-meaningful variability. Latent diffusion, which offers a balance between diffusion models and latent variable approaches, could be investigated to reduce computation time.  

Finally, while architectural refinements and hyperparameter tuning are essential, they are left for future work. This study focuses primarily on the diffusion process, with a flexible backbone that can be easily replaced or upgraded as needed.

\end{document}

%% file: math_commands.tex
\usepackage{amsmath,amsfonts,bm}

\def\eqref#1{equation~\ref{#1}}
\def\1{\bm{1}}

\def\mI{{\bm{I}}}

\DeclareMathAlphabet{\mathsfit}{\encodingdefault}{\sfdefault}{m}{sl}
\SetMathAlphabet{\mathsfit}{bold}{\encodingdefault}{\sfdefault}{bx}{n}

\newcommand{\R}{\mathbb{R}}